\documentclass{article}

\PassOptionsToPackage{numbers, compress}{natbib}

\usepackage[preprint]{neurips_data_2024_arxiv}

\usepackage[utf8]{inputenc} 
\usepackage[T1]{fontenc}    
\usepackage{hyperref}       
\usepackage{url}            
\usepackage{booktabs}       
\usepackage{amsfonts}       
\usepackage{nicefrac}       
\usepackage{microtype}      
\usepackage[dvipsnames]{xcolor}
\usepackage{amsmath}
\usepackage{multirow}
\usepackage[breakable]{tcolorbox}
\usepackage{graphicx}
\usepackage{multicol}
\usepackage{cleveref} 
\usepackage{listings}

\usepackage{mdframed}

\definecolor{cornflowerblue}{rgb}{0.39, 0.58, 0.93}
\hypersetup{
    colorlinks=true,
    linkcolor=cornflowerblue,
    filecolor=magenta,      
    urlcolor=teal,
    citecolor=cornflowerblue,
    pdftitle={GenQA: Generating Millions of Instructions from a Handful of Prompts},
    pdfpagemode=FullScreen,
    }

\title{GenQA: Generating Millions of Instructions from a Handful of Prompts}

\author{
\textbf{Jiuhai Chen, Rifaa Qadri} \\
\textbf{Yuxin Wen, Neel Jain, John Kirchenbauer} \\
\textbf{Tianyi Zhou, Tom Goldstein}\thanks{Correspondence to \texttt{tomg@umd.edu}.\\ Dataset available at \url{https://huggingface.co/datasets/tomg-group-umd/GenQA}.} \\ \\
{University of Maryland}}

\begin{document}

\maketitle

\begin{abstract}
Most public instruction finetuning datasets are relatively small compared to the closed source datasets used to train industry models. To study questions about finetuning at scale, such as curricula and learning rate cooldown schedules, there is a need for industrial-scale datasets. However, this scale necessitates a data generation process that is almost entirely automated. In this work, we study methods for generating large instruction datasets from a single prompt. With little human oversight, we get LLMs to write diverse sets of instruction examples ranging from simple completion tasks to complex multi-turn dialogs across a variety of subject areas. When finetuning a Llama-3 8B base model, our dataset meets or exceeds both WizardLM and Ultrachat on both knowledge-intensive leaderboard tasks as well as conversational evaluations. We release our dataset, the ``generator'' prompts that created it, and our finetuned model checkpoints.

\end{abstract}

\section{Introduction}\label{sec:intro}
Datasets for language model finetuning are typically crafted by hand, crowdsourced from a pool of human annotators, or built by prompting other large language models to edit and augment existing human written datasets \citep{kopf2024openassistant, sharegpt, xu2023wizardlm, mukherjee2023orca,ding2023enhancing}. The requirement of human inputs makes dataset curation an arduous and expensive process.  These costs have also resulted in a split between academic and industrial finetuning practices, with academic datasets comprising hundreds or thousands of samples, and industrial datasets comprising tens of millions.

\begin{figure}[h!]
    \centering
    \includegraphics[width=0.49\linewidth]{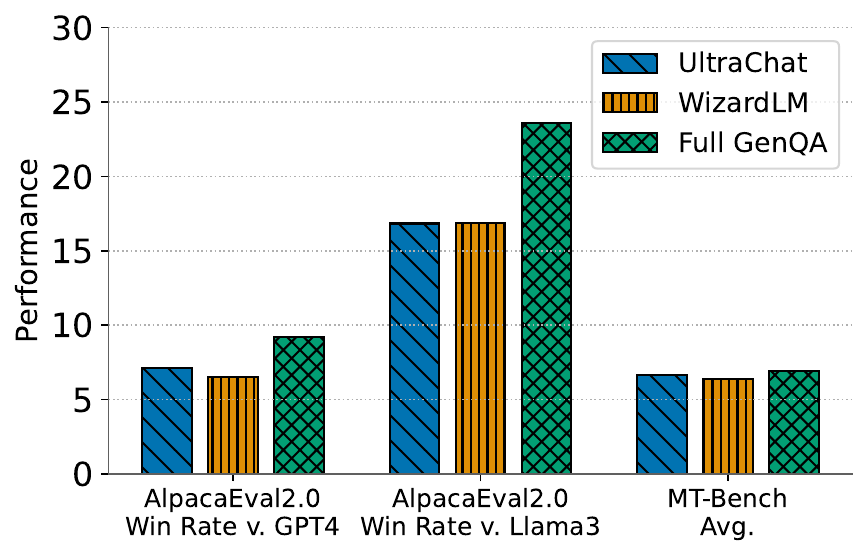}
    \includegraphics[width=0.49\linewidth]{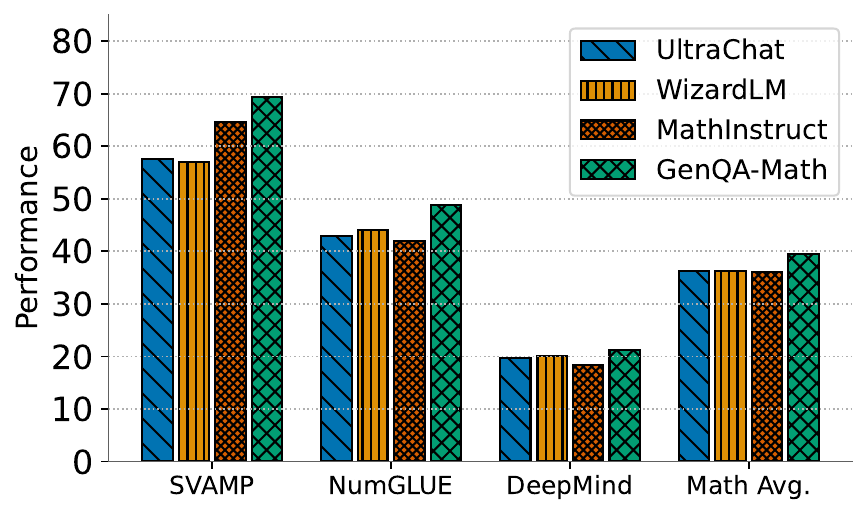}
    \caption{(left) Llama-3-8b finetuned on GenQA performs well on instruction following benchmarks, see \Cref{table:ft-conv}. (right) Finetuning on the math split of GenQA yields strong math performance, see \Cref{table:ft-math}.}    \label{fig:teaser-perf}
\end{figure}

We introduce GenQA, an instruction dataset that is written autonomously by an LLM without being conditioned on human-written questions. We demonstrate that a single hand-written meta-prompt can be used to extract millions of diverse questions from an LLM.  Surprisingly, this automated process can result in high quality datasets that compete with (or even surpass) datasets created using extensive human labor.

Automated dataset creation may seem challenging, as most language models suffer from a lack of randomness. 
 When asked many times to write an instruction/response pair, LLMs may produce a low diversity dataset with many duplicate questions. To extract diverse questions from an LLM in an automated way, we propose {\em generator prompts,} a prompting strategy that boosts the randomness of LLM outputs.  By using a small number of generator prompts, we are able to create a large instruction dataset containing many different question styles across a wide array of subjects.

The GenQA dataset has a few key attributes that make it useful for research purposes.  First, the GenQA dataset natively contains several different splits produced using different kinds of meta prompts.  Second, the GenQA dataset is fairly large, and is meant to reflect the scale of the more than 10M instruction samples reportedly used to fine tune Llama 3. Finally, our empirical evaluation suggests that models trained on GenQA perform well (Figure \ref{fig:teaser-perf}).  In addition to demonstrating the ease with which dataset creation can be automated, we hope the combination of diversity and scale in the GenQA dataset will enable open research on industrial-scale finetuning practices.


\begin{figure}[t!]
    \centering
    \includegraphics[width=0.49\linewidth]{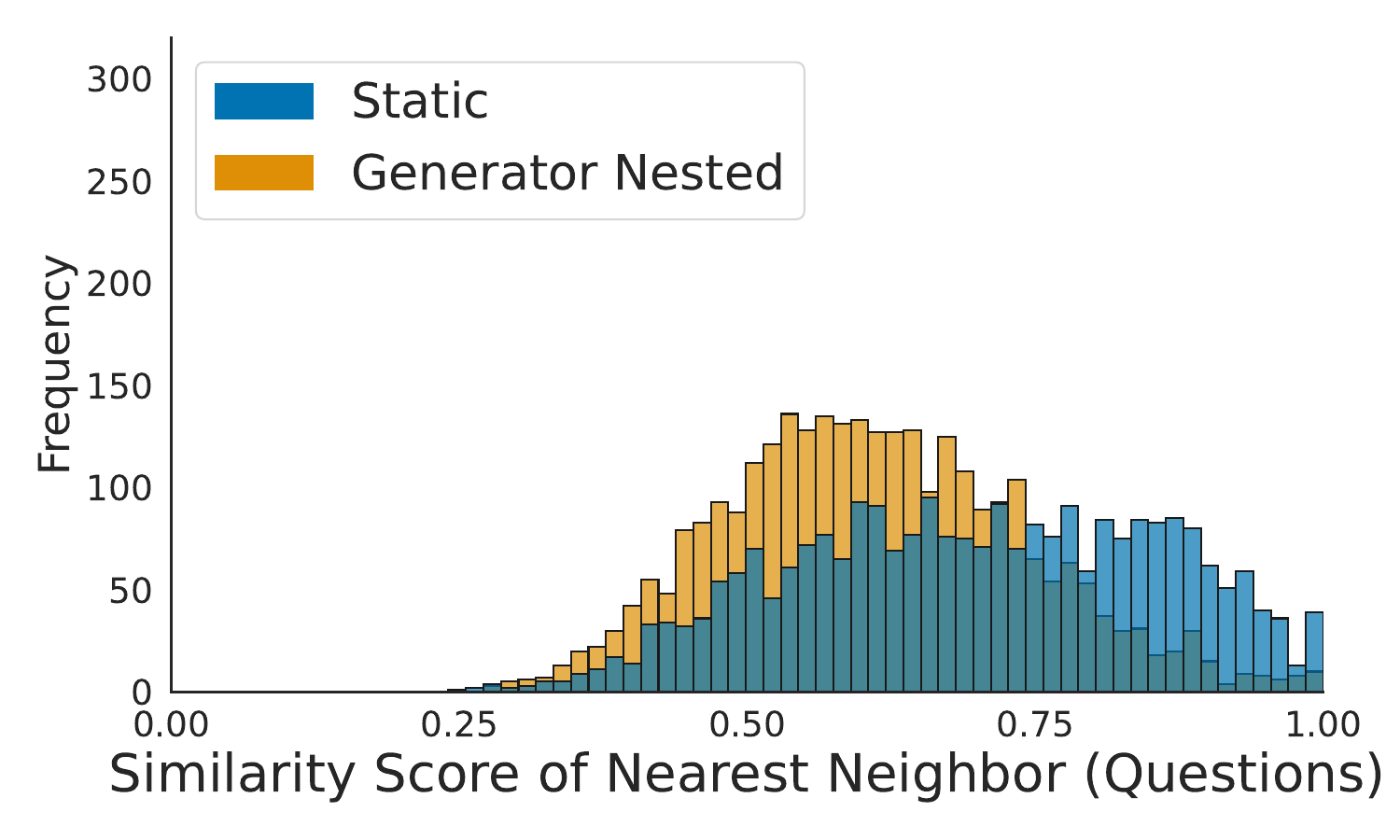}
    \includegraphics[width=0.49\linewidth]{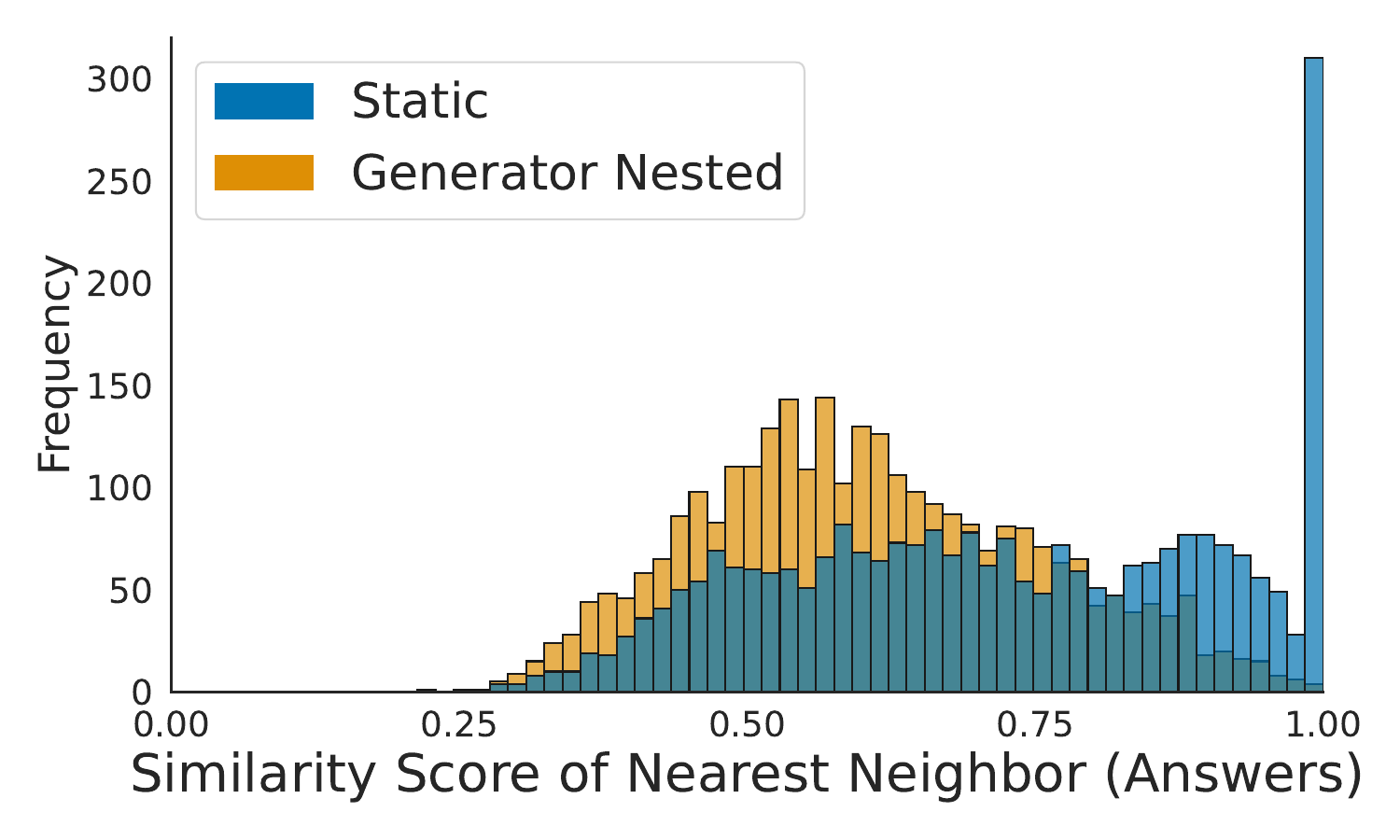}
    \caption{Compared to static prompts that simply ask an LLM for a question/answer pair, our generator prompting strategy results in much higher diversity and more unique questions. Lower similarity scores indicate more diversity, see \Cref{sec:generator-science}.}    \label{fig:teaser-div}
\end{figure}

\section{Related Work}\label{sec:related_work}
One of the earliest forms of modern data for language model finetuning was the Flan collection \citep{wei2021finetuned-FLAN}. 
The multi-task dataset used to train the T0 model family \citep{sanh2021multitask-T0} as well as the work of \citet{FLANchung2022scaling} and \citet{xu2022zeroprompt} expanded these data to include thousands of tasks, achieving further improvements over the original Flan. These datasets, while large and mostly factually correct, suffer from grammatical errors and other major text quality issues.  
\citet{ouyang2022instructgpt} combined multi-task data and a reinforcement learning objective to produce InstructGPT, a finetune of GPT-3 with improved controllability and utility for downstream users. 

Next, \citet{wang2022selfinstruct} showed that an existing finetuned model such as InstructGPT (Text-Davinci-003) could be used to produce instruction-output pairs that were in turn useful to finetune other foundation models in an output based distillation process. One of the most notable demonstrations of this technique was the landmark work that turned the base Llama model \citep{touvron2023llama} into a highly capable instruction-following version, Alpaca, using a dataset of just $52$k outputs sampled from Text-Davinci-003 \citep{alpaca}. As distilled datasets became more popular, the community began constructing others by either filtering and/or augmenting existing datasets in specific ways. \citet{xu2023wizardlm} augmented the original Alpaca dataset \citep{alpaca} to produce \textit{Evol-Instruct}, sometimes referred to as \textit{WizardLM}, using a pipeline that ``evolves'' existing instructions with five types of meta prompts that constrain, deepen, concretize, increase reasoning steps, and generally complicate the original input.

These model augmented instruction datasets were very small, often containing only tens of thousands of examples. To increase the scale at which open source models could be instruction tuned, \citet{mukherjee2023orca} sampled the dataset proposed by \citet{FLANchung2022scaling} and rewrote the responses using ChatGPT and GPT-4 \citep{brown2020language, achiam2023gpt4}. In another approach, \citet{OpenHermes_2_5} and \citet{wang2023far} methodically combined multiple sources of existing instruction tuning datasets to create very large singular datasets. For the domain of mathematics in particular, \citet{yue2023mammoth} compiled many existing mathematical reasoning datasets into a single compendium and further supplemented them using GPT-4 \citep{achiam2023gpt4} to produce MAmmoTH. However, all of these approaches were hampered by the fact that machine generated data tends to lack diversity and concentrate around few modes \citep{zhang2024forcing}. 

In a concurrent work, \citet{xu2024magpie} extract instructions from Llama3 models by prompting them with an empty string, which often results in a random instruction. This same strategy was used to create the {\em general} split of GenQA using GPT-3.5, although the GPT models produce a random answer rather than a random question  (see Appendix \ref{sec:represent}). Our initial experiments with instruction extraction found the empty string approach to be ineffective with commercial models; After making 2M queries to GPT-3.5, the empty string strategy resulted in only 304K unique answers (15\% uniqueness).  We found higher diversity and better control over question topics using our generator prompt strategy, which yielded 85\% unique answers in all of our experiments. 
 
Our approach for building a completely machine generated dataset differs from these prior works both in its scale ($>10\text{M}$ samples) and in our use of specially constructed prompts that result in a set of generated instructions and responses with high diversity in both their structure and the topics that they cover.

\section{The GenQA Dataset}

\begin{figure}
    \centering
    \includegraphics[width=0.49\linewidth]{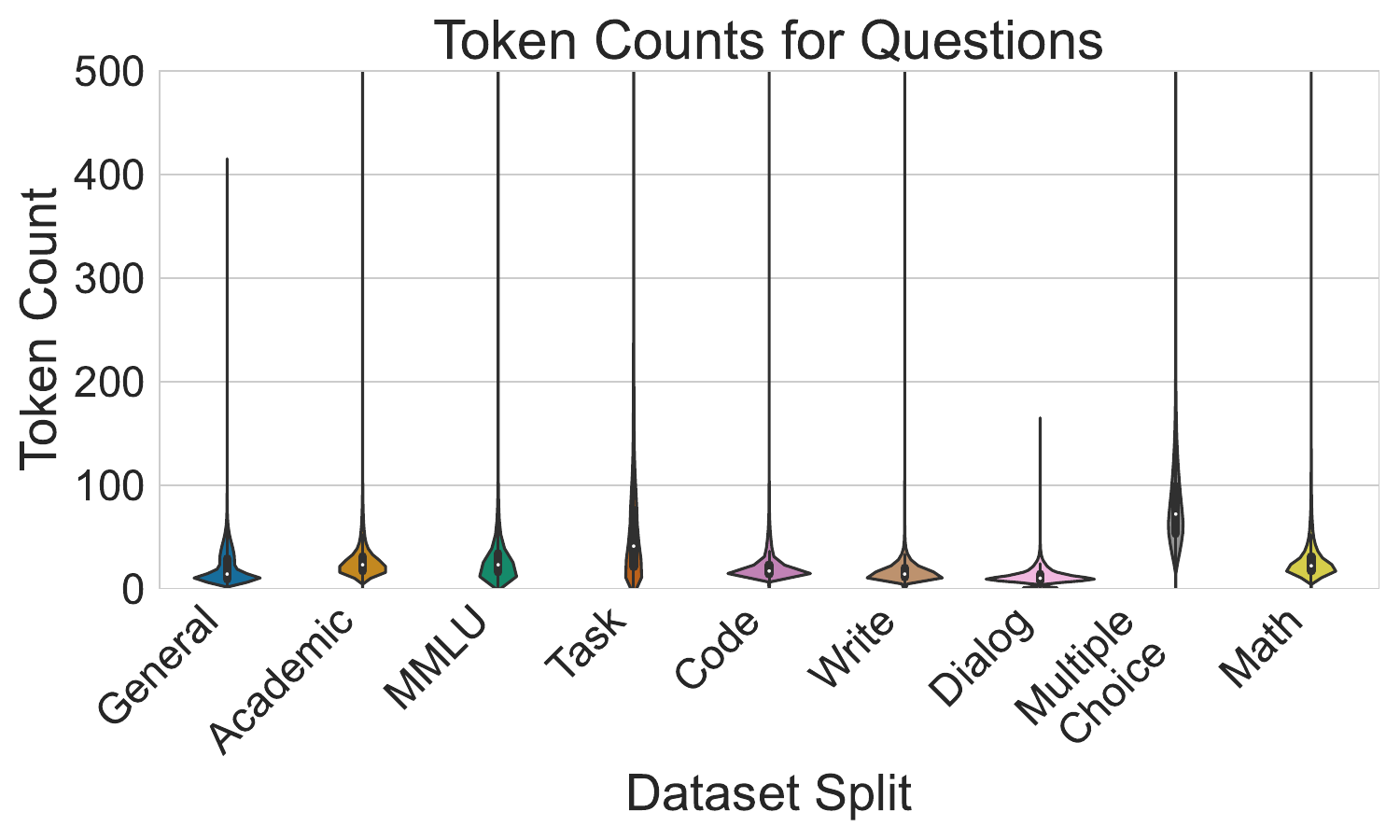}
    \includegraphics[width=0.49\linewidth]{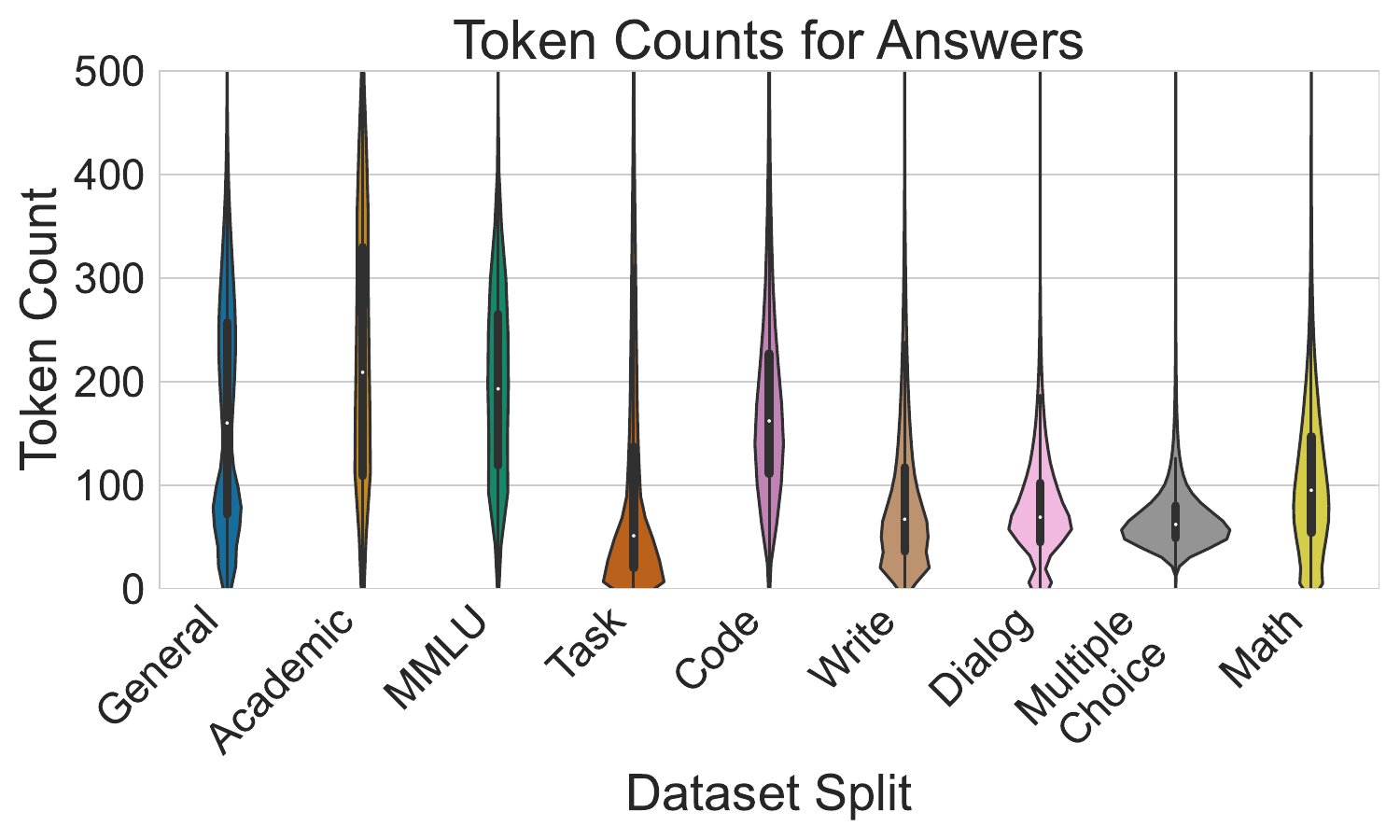}

    \caption{White-space token count for questions (left) and answers (right). Charts are truncated at 500 tokens. Some answers contain over 6K tokens, see \Cref{sec:token-analysis} for the full tail of the distribution.}
    \label{fig:white-space-token-count}
\end{figure}

Table \ref{table:split-sample-counts} lists the splits of GenQA, along with the number of instructions in each and a brief description of the split. The number of tokens in questions and responses is shown in \Cref{fig:white-space-token-count}. See \Cref{sec:token-analysis} for more detailed token counts of multi-turn conversations.
In the main paper below, we describe the methodology used to create the dataset, and give concrete examples of meta-prompts used to create the Academic split. 
\Cref{sec:represent} lists the generator meta-prompt used to create every split, along with a representative question and answer from each split.
All questions were created by Gemini Pro 1.0, with the exception of the General split, which also involved GPT-3.5 (see \Cref{sec:represent} for more details).

In the following sections, we explain how our specially crafted {\em generator prompts} work, and how we used them to construct the GenQA dataset. We then perform a scientific analysis of the dataset to understand which prompting strategies were most effective, and how generator prompts should be designed to promote diversity.  



\begin{table}
\centering
\small
\begin{tabular}{lll}
\toprule
\textbf{Split} & \textbf{Questions} & \textbf{Description} \\
\midrule
Academic & 4,210,076 & QA on a range of academic topics  \\
\midrule
MMLU & 2,409,841 & QA on the topics found in the MMLU dataset \\
\midrule
Multiple Choice & 372,610 & Multiple choice questions on diverse topics \\
\midrule
Writing & 932,362 / 1,864,724 & Compositional writing and editing of documents  \\
\midrule
Task &  1,004,179 / 1,515,280 & Non-compositional text-based tasks\\
\midrule
Code & 513,483 & QA about programming topics in various languages \\
\midrule
Math & 515,509 / 1,104,324 & Math questions, elementary to graduate level \\
\midrule
Dialog & 819,154 / 3,222,818 & Multi-turn conversations containing explanations or advice \\
\midrule
General & 304,920 & QA about pop culture and daily life \\
\bottomrule
\end{tabular}
\vspace{2mm}
\caption{GenQA contains 11,082,134 questions (15,518,076 counting each conversation turn separately) broken into nine splits, each of which were produced using different prompts. In total, the dataset contains approximately 2.8 billion whitespace delimited words.}
\label{table:split-sample-counts}
\end{table}



\subsection{Generator Prompts}\label{sec:generator-methodology}

The quality and diversity of training data are crucial to instruction tuning. Unfortunately, it can be difficult to induce an LLM to produce a large amount of diverse content. Here, we introduce {\em generator prompts}.  We then study the best way to formulate these prompts by quantifying the level of data diversity they yield.

A naive method for automating content generation is simply to choose a topic, and then construct a {\em static prompt} that asks for content on that topic. Consider the toy example of generating a long list of colors. This can be done by feeding the following static prompt to Gemini 1.0 Pro many times:

\texttt{State a random color.  Don't output anything but the color.}

The use of a static prompt yields low diversity. Running the above prompt through the Gemini Pro 1.0 language model $1000$ times produces only $33$ unique outputs.

A {\em generator prompt} boosts diversity by asking a model to produce a long list of possible choices, and then select one of the candidates at random. For example, consider the following prompt:

\texttt{First, print the heading "Colors:", followed by a numbered list of 100 colors.  Then, print the heading "Chosen color:". Then print color number \{N\} on a line by itself.}

The placeholder \texttt{N} should be replaced with a random number each time the prompt is invoked.  When run 1000 times, this prompt yields 383 different colors. One can also use two nested generators as follows.

\texttt{First, print the heading "Colors:", followed by a numbered list of 100 different colors.  Then, print the heading "Chosen color:". Then print color number \{N1\} on a line by itself.  Then, print the heading "Color variants:". Then print a numbered list of 100 different color variants that look like color number \{N1\}, and don't appear on the original "Colors:" list.   Then, print the heading "Chosen variant:". Then print variant number \{N2\} on a line by itself.}

This prompt yields 782 unique colors from 1000 runs.

We hypothesize that the output diversity produced when using a {\em generator prompt} comes from several sources. First, by explicitly creating a list, in the above example we guarantee that there will always be at least 100 candidates, and that these candidates are unique (assuming the LLM followed the provided directions). Second, the process of creating the list requires many sequential samples to be drawn from the model. If the temperature is warm, this compounded randomness makes it highly unlikely that the same list will be produced more than once. 

\subsection{A Study of Generator Prompts for Dataset Creation}
Here, we study how to apply the idea of generator prompts to create random question/answer pairs. We consider several different prompt types in order of increasing complexity. Then, we rigorously study the level of diversity these prompts produce.

Initial attempts to create the Academic split using static prompts resulted in low diversity.  For example, one could choose this prompt:

\textbf{Static: }\texttt{Write a random complex question and its long answer. Begin your question with "Question:" and your answer with "Answer:}

This results in many repeated/identical outputs, motivating us to provide additional context to our prompt. We consider the following prompt conditioned on \texttt{random\_topic}, which is randomly selected from a list of topics written ahead of time by Gemini (see \Cref{sec:topics} for full list).

\textbf{Static-Conditional: }\texttt{Write a complex question from the domain of \{random\_topic\}.  Then write the long answer.  Your question should not contain the words "\{random\_topic\}". Begin your question with "Question:" and your answer with "Answer:"}

To further boost randomness and prevent the model from collapsing into a single mode around each topic, we consider the following generator prompt, also conditioned on \texttt{random\_topic}.

\textbf{Generator-Conditional: }\texttt{List 40 subtopics in the domain of \{random\_topic\}. State subtopic \{N\}.  Then write a question that is not about subtopic \{N\}, but can only be answered with expertise in subtopic \{N\}, and then write the answer. Both the question and answer should be long. The name of the subtopic should not appear in the question.  Begin your questions with "Question:" and your answer with "Answer:".  Be creative.}

Conditioning on a topic prevents the model from collapsing into a small number of modes, but it also constrains the range of possible topics.  In the example shown in~\Cref{sec:generator-methodology}, we see that one can produce randomness using the nested generator approach.  Applying this idea results in the following:

\textbf{Generator-Nested: }\texttt{List 60 topics that you can answer questions about. State topic \{N1\}.  Then write 60 subtopics about topic \{N1\}.  Then state the subtopic \{N2\}.  Then write a question that is not about subtopic \{N2\}, but can only be answered with expertise in subtopic \{N2\}. Then write the answer. Both the question and answer should be long. The name of the subtopic \{N2\} should not appear in the question, and none of the words in subtopic \{N2\} should be reused in the question. Begin your questions with "Question:" and your answer with "Answer:".  Be creative.}


This method has a potential drawback: The LLM sees the selected indices before writing the list, and this may influence the order of the listed items. For this reason, we consider the following construct:

\textbf{Generator-Uniform: }\texttt{List 60 topics that you can answer questions about. Choose a topic uniformly from this list, and state it.  Then write 60 subtopics about the chosen topic.  Then choose a subtopic uniformly from this list, and state it.  Then write a question that is not about the subtopic, but can only be answered with expertise in the subtopic. Then write the answer. Both the question and answer should be long. The name of the subtopic should not appear in the question, and none of the words in subtopic should be reused in the question. Begin your questions with "Question:" and your answer with "Answer:".  Be creative.}

In this construction, the random index is not available to the LLM when the lists are being constructed, as the index it chosen via sampling rather than appearing in the prompt.

Remark: The Gemini model tends to interpret instructions quite literally. If we ask for a question about Cultural Anthropology, we are likely to get a question about Cultural Anthropology {\em per se}, such as ``Who is the father of Cultural Anthropology'' or ``What is that most famous textbook in Cultural Anthropology.'' We avoid this caveat by prompting for a question that is ``not about the subtopic, but can only be answered with expertise in the subtopic.''

\subsection{The Science of Generator Prompts: Which Construction Yields the Most Diversity?}\label{sec:generator-science}



To evaluate these prompting strategies, we generate a separate dataset of size 3000 for each prompt type.\!\footnote{We did exact deduplication on the released the dataset, but this study was performed pre-deduplication.} To measure the level of diversity produced by each prompt, we used the all-MiniLM-L6-v2 retrieval model \cite{wang2020minilm} to encode the first $2$ sentences of each question (or answer) into 384-dimensional dense vectors. For each question (or answer), we record its cosine similarity to its nearest-neighbor. A similarity score of $1.0$ indicates that a question has an exact duplicate, while smaller values indicate that an input is more unique.

\begin{figure}
    \centering
    \includegraphics[width=0.49\linewidth]{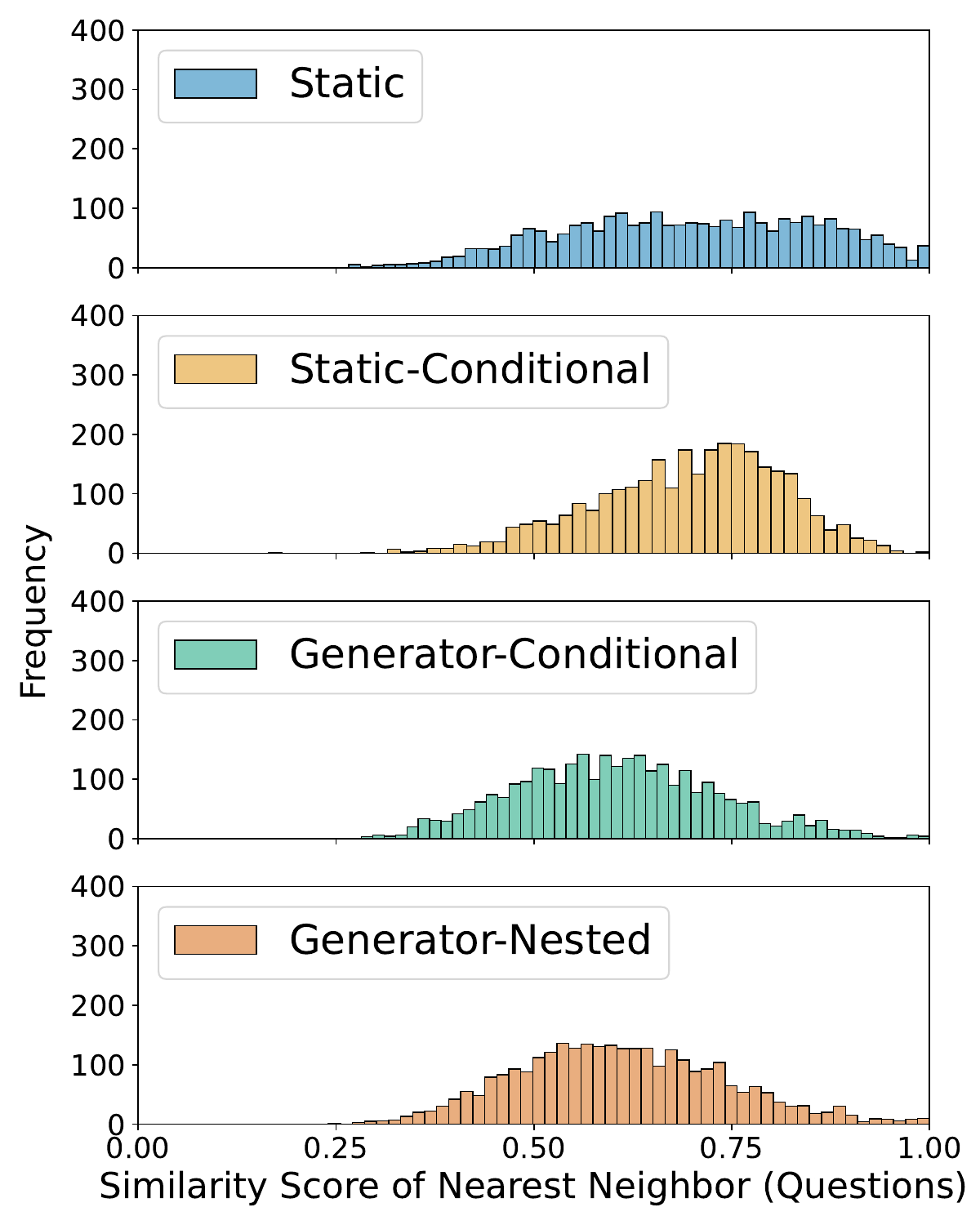}
    \includegraphics[width=0.49\linewidth]{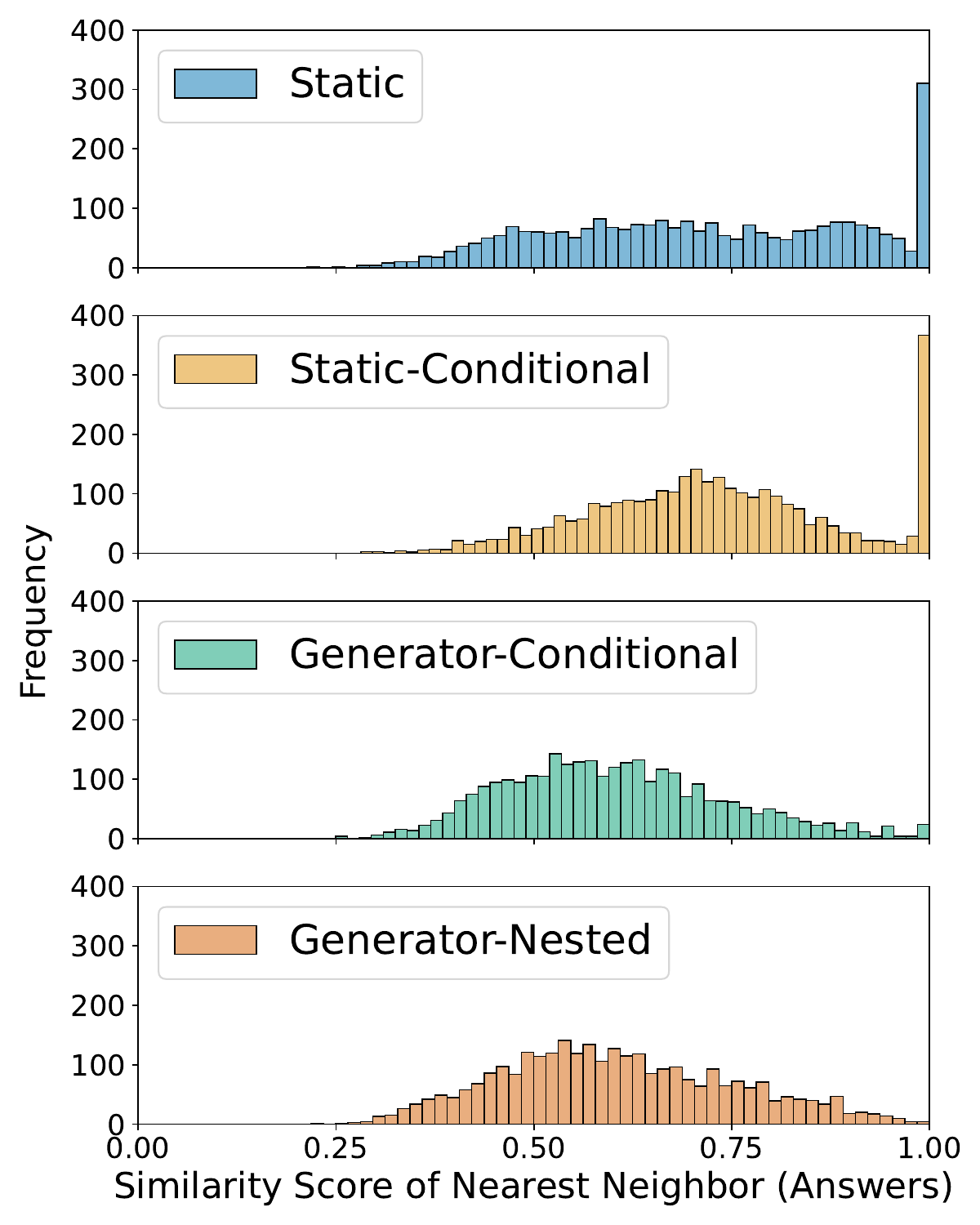}
    \caption{Comparison of nearest-neighbor similarity scores in the Academic split. The generator-conditional and generator-nested strategies perform best.}
    \label{fig:academick1}
\end{figure}

\Cref{fig:academick1} compares the similarity scores within each dataset, computed on either questions or answers. The static prompting strategy resulted in high similarity between nearest neighbors and a higher number of duplicates. The generator-conditional and generator-nested prompts, which were used to create the final {academic} split, yield the highest diversity. Please refer to the appendix figure in \ref{diversity-app} for the full analysis of the prompting strategies evaluated in the Academic split.
%
%
Similar analysis is conducted for the Task split and can also be found in Appendix \ref{diversity-app}. 


\subsection{Adding a Suffix to Boost Randomness}\label{sec:booster-science}
The prompts above end with the phrase ``Be creative.''  We call this a {\em randomness booster}. During the creation of the splits, we randomly append one of the following boosters to the end of the prompt every time it is invoked: ``Be creative,'' ``Be different,'' ``Be smart,'' ``Be weird,'' ``Don't ask the first thing you think of,'' ``Be creative and don't ask the first thing you think of,'' or an empty string (no booster). 

\begin{figure}
    \centering
    \includegraphics[trim={0.0cm 1.0cm 0.0cm 0.0 cm},clip,width=0.75\linewidth]{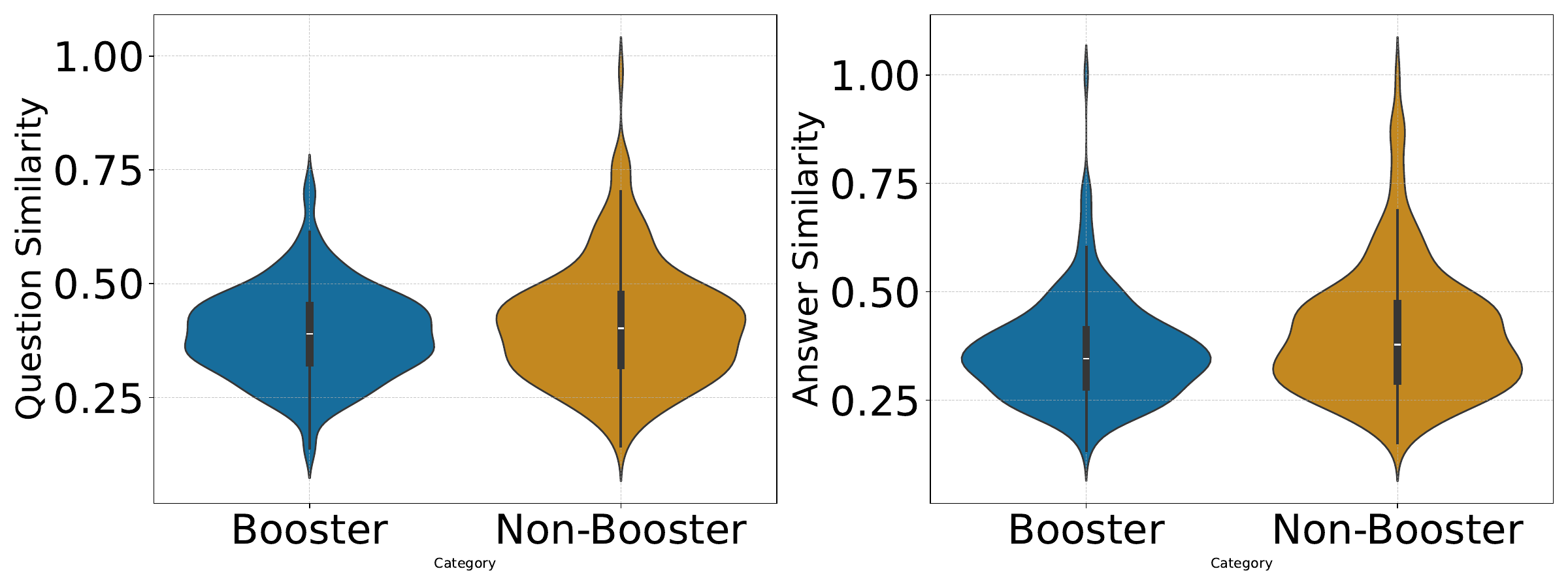}
    \caption{Applying a random booster to the prompts for the Academic split improves data diversity. 
    }
    \label{fig:booster-effect}
    \vspace{-0.25cm}
\end{figure}

To assess the effect of the booster, we sample an equal number of question and answer pairs generated with and without a booster from each split in our dataset ($n=200$ for each type). Following the same procedure in \Cref{sec:generator-science} to analyze diversity, we demonstrate the impact of boosters for the Academic split in \Cref{fig:booster-effect}. The presence of the booster improves diversity across Academic, Task, Multiple Choice, MMLU and Dialogue splits. We include further analysis of boosters in \Cref{sec:app-booster}. 

\subsection{Assembling the Final Dataset}
The GenQA dataset was created by constructing generator prompts, either topic-conditioned or nested, for each split.  Prompts used for each split and example questions generated by the prompts are listed in detail in  
\Cref{sec:represent}.  For each split, the generator prompts were fed through Gemini Pro 1.0 many times and the outputs were parsed into questions and answers.  The final questions were deduplicated using an exact match criteria on the first two sentence of the questions.

\subsection{Comparing Diversity in Other Finetuning Datasets}\label{sec:other-ds-science}

\begin{figure}
    \centering
    \includegraphics[trim={0.0cm 0.8cm 0.0cm 0.0cm},clip,width=0.45\linewidth]{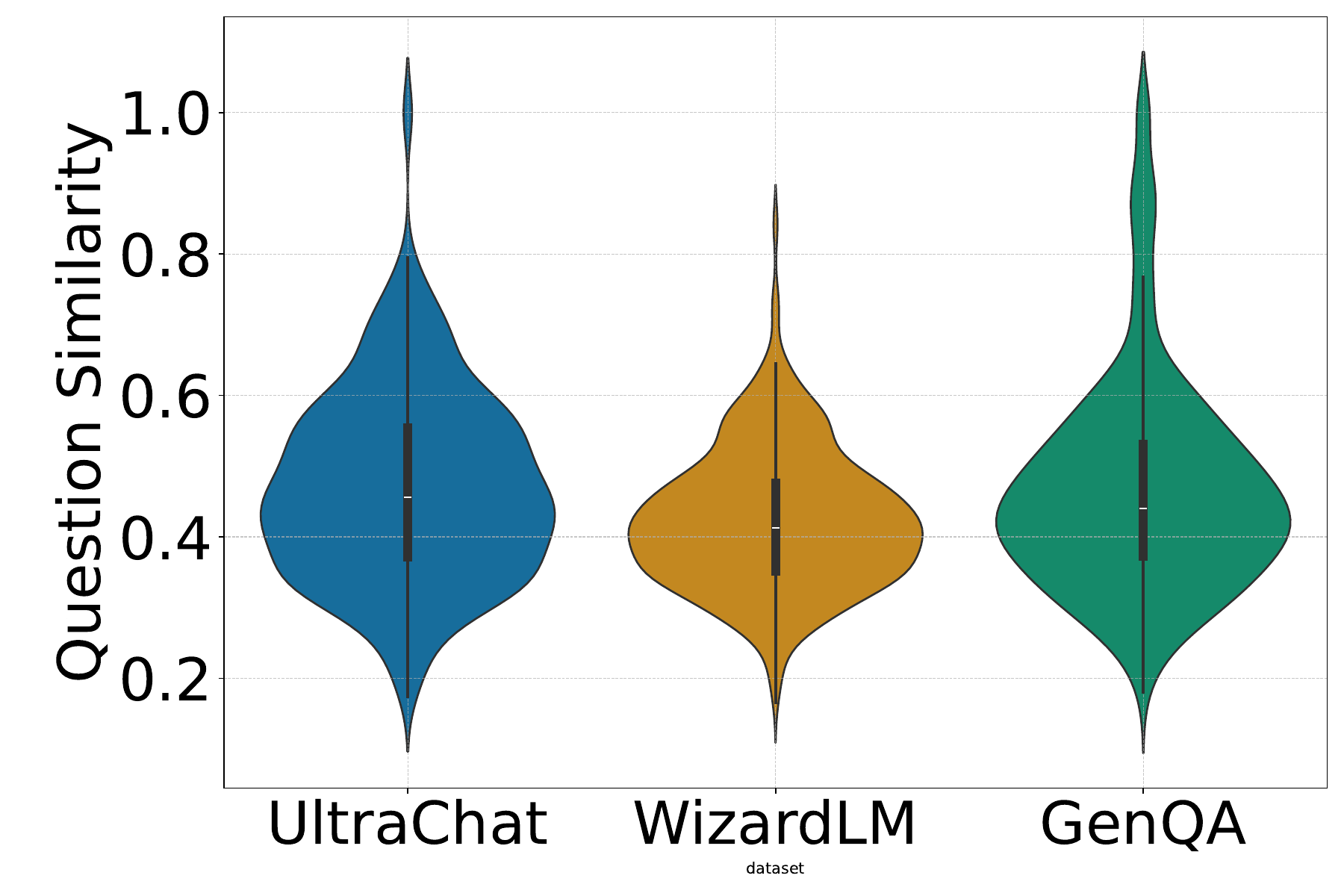}
    \includegraphics[trim={0.0cm 0.8cm 0.0cm 0.0cm},clip,width=0.45\linewidth]{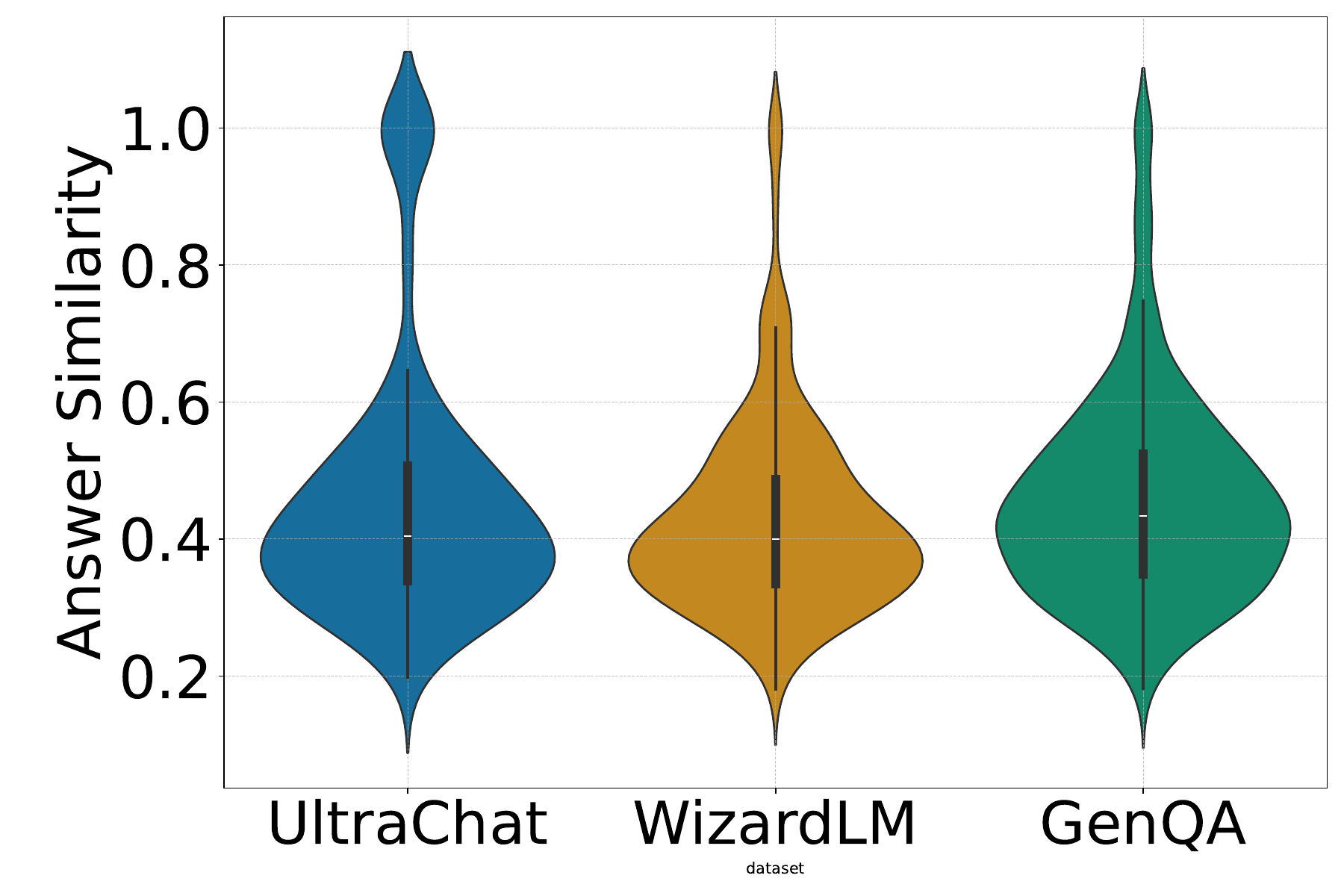}
    \caption{Comparison of nearest neighbor ($k=1$) similarity across GenQA, UltraChat, and WizardLM.}
    \label{fig:other-ds-diversity-violin}
\end{figure}

We compare the diversity of GenQA to existing finetuning datasets {\em WizardLM} and {\em UltraChat} by uniformly sampling equal sized subsets from each dataset and analyzing their diversity in the same manner as \Cref{sec:generator-science}. We observe that the diversity of GenQA is on-par with that of the reference datasets, as seen in \Cref{fig:other-ds-diversity-violin}.

\section{Finetuning Experiments}\label{sec:finetuning}

To demonstrate the quality of GenQA for language model finetuning, we perform an empirical evaluation against other strong finetuning datasets.

\subsection{Finetuning Setup}\label{sec:ft-setup}
We tune a Llama-3-8B \citep{llama3modelcard} model with the default chat template on GenQA and two existing instruction finetuning datasets:  UltraChat \citep{ding2023enhancing} and WizardLM \citep{xu2023wizardlm}. To evaluate the final models we consider tasks from the Huggingface Open LLM Leaderboard and two instruction-following benchmarks. We use the Hugging Face Alignment Handbook codebase \cite{alignment_handbook2023} for our finetuning runs and the same set of standard hyperparameters for each dataset. Full details are provided in \Cref{sec:ft-hparams}. The baseline instruction datasets and evaluation benchmarks we consider are described below.

{\em WizardLM-Evol-Instruct-V2} \citep{xu2023wizardlm} contains $196$k single-turn instructions. The dataset is developed by starting with initial instructions from a base dataset such as Alpaca \citep{alpaca}. It is then enhanced using a large language model like GPT-4, which incrementally increases the complexity of the instructions through various strategies\footnote{\url{https://huggingface.co/datasets/MaziyarPanahi/WizardLM_evol_instruct_V2_196k}}.

{\em UltraChat} \citep{ding2023enhancing} is another synthetic instruction dataset, specifically focusing on multi-turn conversational abilities. The dataset is generated by simulating conversations between two large language models on three different topics: general questions, writing, and assistance, aiming to ensure diversity. In this paper, we use a filtered version of UltraChat with a total of $200$k multi-turn instructions \footnote{\url{https://huggingface.co/datasets/HuggingFaceH4/ultrachat_200k}}.

\textbf{Evaluation} To demonstrate the capabilities of our finetuned models, we evaluate them on a variety of general benchmark tasks and conversational benchmarks. For general benchmark tasks, we include ARC \citep{Clark2018ThinkYH}, BoolQ \citep{clark2019boolq}, HellaSwag \citep{zellers2019hellaswag}, MMLU \citep{hendrycks2020measuring}, OpenBookQA \citep{OpenBookQA2018}, PIQA \citep{Bisk2020}, TruthfulQA \citep{lin2021truthfulqa}, and Winogrande \citep{sakaguchi2019winogrande}. This diverse range of benchmarks assesses the models' reasoning, knowledge, and truthfulness. Additionally, we report scores for AlpacaEval 2.0 length-control and MT-bench. These benchmarks test the models' conversational abilities and their capacity to follow instructions.

\textbf{Rebalancing the GenQA Splits} The GenQA dataset is not balanced across splits, with the Academic split comprising $\sim 38\%$ of the entire dataset in its raw form. We find that we get best finetuning performance using adjusted sampling ratios that up-weight the smaller splits. See \Cref{tab:rebalance-ratios} in \Cref{sec:ft-hparams}. We refer to the rebalanced version of the dataset as ``Full GenQA'' and the ``Subset GenQA'' versions at various token counts are a further random sample of the rebalanced dataset.

\subsection{Results of Finetuning}

\begin{table}[t!]
\setlength{\tabcolsep}{2.5pt}
\centering
\caption{Performance on two different benchmarks that measure a model's ability to engage in coherent, informative, and engaging conversations, AlpacaEval and MT-Bench.}
\label{table:ft-conv}
\begin{tabular}{c|ccc|ccc}
\toprule
\multirow{2}{*}{Dataset} & \multicolumn{3}{c|}{AlpacaEval 2.0} & \multicolumn{3}{c}{MT-bench} \\
  & v.s. GPT4 & v.s. Llama-3-8B-Instruct & Length & 1-round & 2-round & Avg.\\
\midrule
WizardLM  & 7.11         & 16.84       & 1522      & 7.18      & 6.12  & 6.65\\
Subset GenQA (135M tokens) & 7.75         & 17.49       & 1498                 & 7.26   & 6.06 & 6.67 \\
\midrule
UltraChat  & 6.50         & 16.88           & 1282             & 6.93      & 5.88  & 6.40 \\
Subset GenQA (238M tokens)   & \textbf{9.31}         & 20.41   & 1096                     & 7.49    & 6.15  & 6.82  \\
\midrule
Full GenQA    & 9.20                       & \textbf{23.57} &  \textbf{1060}   & \textbf{7.55} & \textbf{6.26} & \textbf{6.91}  \\  
\bottomrule
\end{tabular}
\end{table}


\begin{table}[t!]
\scriptsize
\setlength{\tabcolsep}{3.5pt}
\centering
\caption{Performance on various reasoning, knowledge, and truthfulness benchmarks. GenQA is on par with the other models, including Llama-3-8B-Instruct.}
\label{table:ft-leader}
\begin{tabular}{ccccccccccc}
\toprule
   & ARC\_E & ARC\_C & \multicolumn{1}{l}{BoolQ} & \multicolumn{1}{l}{HellaSwag} & \multicolumn{1}{l}{MMLU} & \multicolumn{1}{l}{OpenBookQA} & \multicolumn{1}{l}{PIQA} & \multicolumn{1}{l}{TruthfulQA} & \multicolumn{1}{l}{Winogrande} & \multicolumn{1}{l}{Avg} \\
\midrule
Llama-3-8B   & 52.99 & 77.86 & 80.95 & \textbf{79.15} &	62.01 &	34.80 &	80.79 &	43.80 & 73.56 & 65.10\\
Llama-3-8B-Instruct   & \textbf{56.91} &	79.67 &	83.15 &	75.79 &	\textbf{63.81} &	42.80 &	78.73 &	51.70 &	71.67 &	67.14\\
\midrule
WizardLM   & 55.12 & 79.63 & 82.02 &	78.38 &	61.42 &	45.40 &	80.79 &	\textbf{53.03} &	\textbf{74.59} &	\textbf{67.82}  \\
\begin{tabular}[c]{@{}c@{}}Subset GenQA\\ (135M tokens) \end{tabular} & 55.46	& \textbf{80.30}	& 83.73	& 78.65	& 60.88	& \textbf{46.00} & 81.18 & 48.16 & 74.51 &	67.65      \\
\midrule
UltraChat  & 54.86 &	79.76 &	83.39 &	78.60 &	62.05 &	44.20 &	\textbf{81.45} &	49.05 &	73.88 &	67.47 \\
\begin{tabular}[c]{@{}c@{}}Subset GenQA\\ (238M tokens) \end{tabular} & 55.38 & 80.22 & \textbf{83.82} & 78.51 & 60.51 & 45.60 & 81.18 & 48.26 & 74.11 &	67.51   \\
\midrule
Full GenQA  & 55.46 &	80.13 &	83.70 &	78.81 &	61.07 &	\textbf{46.00} &	81.28 &	47.06 &	74.03 &	 67.50\\
\bottomrule
\end{tabular}
\end{table}

We showcase the results of finetuning on different datasets in \Cref{table:ft-conv}. The model finetuned on GenQA achieves the highest scores on both AlpacaEval and MT-bench. Additionally, we report the results of Leaderboard tasks in \Cref{table:ft-leader}. The model finetuned on GenQA achieves comparable results to those of models finetuned on the baseline datasets. Overall we find that GenQA is performant when evaluated from a token-for-token sample complexity perspective against other datasets, and we also observe even further improvement if we leverage its size by training on its totality.

\paragraph{Token-for-Token}
To ensure a fair comparison with GenQA, given its significantly larger scale than the WizardLM and UltraChat datasets, we randomly sample a subset from GenQA to match the token count of the baseline datasets. In \Cref{table:ft-conv} and \Cref{table:ft-leader}, we present the results of the model finetuned on a subset of GenQA, referred to as ``Subset GenQA,'' which has the same number of tokens as the baseline dataset it is paired with.

The token-for-token comparison in \Cref{table:ft-conv} reveals that the model finetuned on GenQA outperforms the UltraChat dataset according to both instruction-following benchmarks evaluated and also is comparable to tuning on the WizardLM dataset.
Measured on the series of knowledge and reasoning leaderboard benchmarks tabulated in \Cref{table:ft-leader}, GenQA is also competitive. We find that measured across all benchmarks in aggregate, and controlling for token count, the average performance of GenQA slightly beats UltraChat and slightly underperforms WizardLM, but is generally comparable\footnote{To calibrate expectations, we observe that finetuning on any of the instruction datasets achieves results comparable to the official Llama-3-8B-Instruct model, surpassing the base model Llama-3-8B handily.}. 

In summary, our token-for-token comparison indicates that despite being \textit{totally machine generated} from a handful of prompts without conditioning on human written questions, GenQA yields high-quality finetuned models. The baseline datasets we compare it to are derivatives of existing data, requiring further calls to state of the art models like GPT-4 during their augmentation process, and/or must undergo additional data curation steps to enhance their complexity. 

\paragraph{Training on Individual Splits}
We also conduct a series of experiments where we finetune the model exclusively on the Math split of GenQA. This allows us to compare it to other large expert datasets like MathInstruct \citep{yue2023mammoth}, which is specifically designed to enhance Mathematical reasoning. We evaluate the finetuned models on various Mathematical reasoning datasets including Math \citep{hendrycks2021measuring}, GSM8K \citep{cobbe2021training}, SVAMP \citep{patel2021nlp}, NumGLUE \citep{mishra2022numglue}, DeepMind \citep{saxton2019analysing}, and SimulEq \citep{Mishra2022Lila}. As shown in \Cref{table:ft-math}, a model finetuned on the GenQA Math subset outperforms the baselines in all benchmarks except SimulEq, where the models finetuned on WizardLM and UltraChat perform marginally better.

\begin{table}[t!]
\centering
\caption{Performance on various Math reasoning tasks with 5 shots Chain-of-Thought (CoT) prompting. The Math split of GenQA outperforms all other datasets including a similarly sized but random subset of GenQA, and a Math specific instruction tuning dataset, MathInstruct \citep{yue2023mammoth}.}
\label{table:ft-math}
\begin{tabular}{cccccccc}
\toprule
   & Math & GSM8K & SVAMP & NumGLUE & DeepMind & SimulEq & \multicolumn{1}{l}{Avg.} \\
\midrule
WizardLM & 17.10 & \textbf{58.07} & 57.60 &	42.90 & 19.70 & \textbf{21.98} & 36.23\\
\begin{tabular}[c]{@{}c@{}}Subset GenQA\\ (125M tokens) \end{tabular} & 18.68 & 56.56 &	65.00 & 46.64 & 19.90 & 19.26 & 37.67 \\
\midrule
UltraChat & 18.06 & 57.85 & 57.00 & 44.05 &20.10 & 20.43 & 36.25\\
\begin{tabular}[c]{@{}c@{}}Subset GenQA\\ (238M tokens) \end{tabular}	& 18.66	& 55.72 & 68.90 & 47.41 & 19.30 & 20.04 & 38.33 \\
\midrule
\begin{tabular}[c]{@{}c@{}}MathInstruct \\ (114M tokens) \end{tabular}   & 19.56 &	56.79 &	64.70 & 42.03 & 18.30 & 14.98 &	36.06    \\
\midrule
\begin{tabular}[c]{@{}c@{}}GenQA-Math\\ (222M tokens) \end{tabular}  & \textbf{20.30} & 56.86 & \textbf{69.40} &	\textbf{48.85} &	\textbf{21.20} & 20.23 & \textbf{39.47}  \\
\bottomrule
\end{tabular}
\end{table}

\paragraph{Full GenQA: Is bigger better?}\label{sec:bigger-better}

We observe that under the instruction-following evaluation, the model trained on all of GenQA performs better than the models trained only on WizardLM or UltraChat, or the subset of GenQA, in nearly all cases. Furthermore, on the Leaderboard tasks for which GenQA outperforms its peers, extending training from the subset to the full GenQA further improved scores.

It should be noted, however, that on Leaderboard tasks where the subset of GenQA did not yield better performance than the baselines, continued training was not able to make up the difference. This suggests that GenQA still has some blindspots, and like other popular datasets it may benefit from being mixed into a larger cocktail
\citep{OpenHermes_2_5}.

\section{Conclusion}\label{sec:conclusion}

GenQA is an instruction dataset written autonomously by an LLM without conditioning on human questions or using complex multi-stage pipelines. Beyond the obvious uses of the public GenQA samples to improve performance of open source models, we hope that the  methods in this paper can serve as a Swiss army knife for easily creating datasets for other domains.  Our experiments indicate that prompt engineering alone can yield millions of diverse training samples with quality as good as (or in some cases surpassing) high-cost human labelling. The generator prompt strategy can be used to quickly generate datasets anew, or to create data to cover the blindspots of other existing data sources.

\section*{Acknowledgements}
Financial support was provided by the ONR MURI program and the AFOSR MURI program. Private support was provided by Capital One Bank, the Amazon Research Award program, and Open Philanthropy. Further support was provided by the National Science Foundation (IIS-2212182), and by the NSF TRAILS Institute (2229885). Compute support was provided through the Department of Energy INCITE Allocation Program. 


\bibliographystyle{plainnat}
\bibliography{main.bib}

\begin{thebibliography}{38}
\providecommand{\natexlab}[1]{#1}
\providecommand{\url}[1]{\texttt{#1}}
\expandafter\ifx\csname urlstyle\endcsname\relax
  \providecommand{\doi}[1]{doi: #1}\else
  \providecommand{\doi}{doi: \begingroup \urlstyle{rm}\Url}\fi

\bibitem[Achiam et~al.(2023)Achiam, Adler, Agarwal, Ahmad, Akkaya, Aleman, Almeida, Altenschmidt, Altman, Anadkat, et~al.]{achiam2023gpt4}
Josh Achiam, Steven Adler, Sandhini Agarwal, Lama Ahmad, Ilge Akkaya, Florencia~Leoni Aleman, Diogo Almeida, Janko Altenschmidt, Sam Altman, Shyamal Anadkat, et~al.
\newblock Gpt-4 technical report.
\newblock \emph{arXiv preprint arXiv:2303.08774}, 2023.

\bibitem[AI@Meta(2024)]{llama3modelcard}
AI@Meta.
\newblock Llama 3 model card.
\newblock 2024.
\newblock URL \url{https://github.com/meta-llama/llama3/blob/main/MODEL_CARD.md}.

\bibitem[Bisk et~al.(2020)Bisk, Zellers, Bras, Gao, and Choi]{Bisk2020}
Yonatan Bisk, Rowan Zellers, Ronan~Le Bras, Jianfeng Gao, and Yejin Choi.
\newblock Piqa: Reasoning about physical commonsense in natural language.
\newblock In \emph{Thirty-Fourth AAAI Conference on Artificial Intelligence}, 2020.

\bibitem[Brown et~al.(2020)Brown, Mann, Ryder, Subbiah, Kaplan, Dhariwal, Neelakantan, Shyam, Sastry, Askell, et~al.]{brown2020language}
Tom Brown, Benjamin Mann, Nick Ryder, Melanie Subbiah, Jared~D Kaplan, Prafulla Dhariwal, Arvind Neelakantan, Pranav Shyam, Girish Sastry, Amanda Askell, et~al.
\newblock Language models are few-shot learners.
\newblock \emph{Advances in neural information processing systems}, 33:\penalty0 1877--1901, 2020.

\bibitem[Chung et~al.(2022)Chung, Hou, Longpre, Zoph, Tay, Fedus, Li, Wang, Dehghani, Brahma, et~al.]{FLANchung2022scaling}
Hyung~Won Chung, Le~Hou, Shayne Longpre, Barret Zoph, Yi~Tay, William Fedus, Eric Li, Xuezhi Wang, Mostafa Dehghani, Siddhartha Brahma, et~al.
\newblock Scaling instruction-finetuned language models.
\newblock \emph{arXiv preprint arXiv:2210.11416}, 2022.

\bibitem[Clark et~al.(2019)Clark, Lee, Chang, Kwiatkowski, Collins, and Toutanova]{clark2019boolq}
Christopher Clark, Kenton Lee, Ming-Wei Chang, Tom Kwiatkowski, Michael Collins, and Kristina Toutanova.
\newblock Boolq: Exploring the surprising difficulty of natural yes/no questions.
\newblock \emph{arXiv preprint arXiv:1905.10044}, 2019.

\bibitem[Clark et~al.(2018)Clark, Cowhey, Etzioni, Khot, Sabharwal, Schoenick, and Tafjord]{Clark2018ThinkYH}
Peter Clark, Isaac Cowhey, Oren Etzioni, Tushar Khot, Ashish Sabharwal, Carissa Schoenick, and Oyvind Tafjord.
\newblock Think you have solved question answering? try arc, the ai2 reasoning challenge.
\newblock \emph{ArXiv}, abs/1803.05457, 2018.

\bibitem[Cobbe et~al.(2021)Cobbe, Kosaraju, Bavarian, Chen, Jun, Kaiser, Plappert, Tworek, Hilton, Nakano, et~al.]{cobbe2021training}
Karl Cobbe, Vineet Kosaraju, Mohammad Bavarian, Mark Chen, Heewoo Jun, Lukasz Kaiser, Matthias Plappert, Jerry Tworek, Jacob Hilton, Reiichiro Nakano, et~al.
\newblock Training verifiers to solve math word problems.
\newblock \emph{arXiv preprint arXiv:2110.14168}, 2021.

\bibitem[Ding et~al.(2023)Ding, Chen, Xu, Qin, Zheng, Hu, Liu, Sun, and Zhou]{ding2023enhancing}
Ning Ding, Yulin Chen, Bokai Xu, Yujia Qin, Zhi Zheng, Shengding Hu, Zhiyuan Liu, Maosong Sun, and Bowen Zhou.
\newblock Enhancing chat language models by scaling high-quality instructional conversations.
\newblock \emph{arXiv preprint arXiv:2305.14233}, 2023.

\bibitem[Hendrycks et~al.(2020)Hendrycks, Burns, Basart, Zou, Mazeika, Song, and Steinhardt]{hendrycks2020measuring}
Dan Hendrycks, Collin Burns, Steven Basart, Andy Zou, Mantas Mazeika, Dawn Song, and Jacob Steinhardt.
\newblock Measuring massive multitask language understanding.
\newblock \emph{arXiv preprint arXiv:2009.03300}, 2020.

\bibitem[Hendrycks et~al.(2021)Hendrycks, Burns, Kadavath, Arora, Basart, Tang, Song, and Steinhardt]{hendrycks2021measuring}
Dan Hendrycks, Collin Burns, Saurav Kadavath, Akul Arora, Steven Basart, Eric Tang, Dawn Song, and Jacob Steinhardt.
\newblock Measuring mathematical problem solving with the math dataset.
\newblock \emph{arXiv preprint arXiv:2103.03874}, 2021.

\bibitem[K{\"o}pf et~al.(2024)K{\"o}pf, Kilcher, von R{\"u}tte, Anagnostidis, Tam, Stevens, Barhoum, Nguyen, Stanley, Nagyfi, et~al.]{kopf2024openassistant}
Andreas K{\"o}pf, Yannic Kilcher, Dimitri von R{\"u}tte, Sotiris Anagnostidis, Zhi~Rui Tam, Keith Stevens, Abdullah Barhoum, Duc Nguyen, Oliver Stanley, Rich{\'a}rd Nagyfi, et~al.
\newblock Openassistant conversations-democratizing large language model alignment.
\newblock \emph{Advances in Neural Information Processing Systems}, 36, 2024.

\bibitem[Lin et~al.(2021)Lin, Hilton, and Evans]{lin2021truthfulqa}
Stephanie Lin, Jacob Hilton, and Owain Evans.
\newblock Truthfulqa: Measuring how models mimic human falsehoods.
\newblock \emph{arXiv preprint arXiv:2109.07958}, 2021.

\bibitem[Loshchilov and Hutter(2017)]{loshchilov2017decoupled}
Ilya Loshchilov and Frank Hutter.
\newblock Decoupled weight decay regularization.
\newblock \emph{arXiv preprint arXiv:1711.05101}, 2017.

\bibitem[Mihaylov et~al.(2018)Mihaylov, Clark, Khot, and Sabharwal]{OpenBookQA2018}
Todor Mihaylov, Peter Clark, Tushar Khot, and Ashish Sabharwal.
\newblock Can a suit of armor conduct electricity? a new dataset for open book question answering.
\newblock In \emph{EMNLP}, 2018.

\bibitem[Mishra et~al.(2022{\natexlab{a}})Mishra, Finlayson, Lu, Tang, Welleck, Baral, Rajpurohit, Tafjord, Sabharwal, Clark, and Kalyan]{Mishra2022Lila}
Swaroop Mishra, Matthew Finlayson, Pan Lu, Leonard Tang, Sean Welleck, Chitta Baral, Tanmay Rajpurohit, Oyvind Tafjord, Ashish Sabharwal, Peter Clark, and Ashwin Kalyan.
\newblock Lila: A unified benchmark for mathematical reasoning.
\newblock In \emph{Proceedings of the 2022 Conference on Empirical Methods in Natural Language Processing (EMNLP)}, 2022{\natexlab{a}}.

\bibitem[Mishra et~al.(2022{\natexlab{b}})Mishra, Mitra, Varshney, Sachdeva, Clark, Baral, and Kalyan]{mishra2022numglue}
Swaroop Mishra, Arindam Mitra, Neeraj Varshney, Bhavdeep Sachdeva, Peter Clark, Chitta Baral, and Ashwin Kalyan.
\newblock Numglue: A suite of fundamental yet challenging mathematical reasoning tasks.
\newblock \emph{arXiv preprint arXiv:2204.05660}, 2022{\natexlab{b}}.

\bibitem[Mukherjee et~al.(2023)Mukherjee, Mitra, Jawahar, Agarwal, Palangi, and Awadallah]{mukherjee2023orca}
Subhabrata Mukherjee, Arindam Mitra, Ganesh Jawahar, Sahaj Agarwal, Hamid Palangi, and Ahmed Awadallah.
\newblock Orca: Progressive learning from complex explanation traces of gpt-4.
\newblock \emph{arXiv preprint arXiv:2306.02707}, 2023.

\bibitem[Ouyang et~al.(2022)Ouyang, Wu, Jiang, Almeida, Wainwright, Mishkin, Zhang, Agarwal, Slama, Ray, Schulman, Hilton, Kelton, Miller, Simens, Askell, Welinder, Christiano, Leike, and Lowe]{ouyang2022instructgpt}
Long Ouyang, Jeffrey Wu, Xu~Jiang, Diogo Almeida, Carroll Wainwright, Pamela Mishkin, Chong Zhang, Sandhini Agarwal, Katarina Slama, Alex Ray, John Schulman, Jacob Hilton, Fraser Kelton, Luke Miller, Maddie Simens, Amanda Askell, Peter Welinder, Paul~F Christiano, Jan Leike, and Ryan Lowe.
\newblock Training language models to follow instructions with human feedback.
\newblock In S.~Koyejo, S.~Mohamed, A.~Agarwal, D.~Belgrave, K.~Cho, and A.~Oh, editors, \emph{Advances in Neural Information Processing Systems}, volume~35, pages 27730--27744. Curran Associates, Inc., 2022.
\newblock URL \url{https://proceedings.neurips.cc/paper_files/paper/2022/file/b1efde53be364a73914f58805a001731-Paper-Conference.pdf}.

\bibitem[Patel et~al.(2021)Patel, Bhattamishra, and Goyal]{patel2021nlp}
Arkil Patel, Satwik Bhattamishra, and Navin Goyal.
\newblock Are nlp models really able to solve simple math word problems?
\newblock \emph{arXiv preprint arXiv:2103.07191}, 2021.

\bibitem[Sakaguchi et~al.(2019)Sakaguchi, Bras, Bhagavatula, and Choi]{sakaguchi2019winogrande}
Keisuke Sakaguchi, Ronan~Le Bras, Chandra Bhagavatula, and Yejin Choi.
\newblock Winogrande: An adversarial winograd schema challenge at scale.
\newblock \emph{arXiv preprint arXiv:1907.10641}, 2019.

\bibitem[Sanh et~al.(2021)Sanh, Webson, Raffel, Bach, Sutawika, Alyafeai, Chaffin, Stiegler, Raja, Dey, et~al.]{sanh2021multitask-T0}
Victor Sanh, Albert Webson, Colin Raffel, Stephen Bach, Lintang Sutawika, Zaid Alyafeai, Antoine Chaffin, Arnaud Stiegler, Arun Raja, Manan Dey, et~al.
\newblock Multitask prompted training enables zero-shot task generalization.
\newblock In \emph{International Conference on Learning Representations}, 2021.

\bibitem[Saxton et~al.(2019)Saxton, Grefenstette, Hill, and Kohli]{saxton2019analysing}
David Saxton, Edward Grefenstette, Felix Hill, and Pushmeet Kohli.
\newblock Analysing mathematical reasoning abilities of neural models.
\newblock \emph{arXiv preprint arXiv:1904.01557}, 2019.

\bibitem[ShareGPT(2023)]{sharegpt}
ShareGPT, 2023.
\newblock URL \url{https://sharegpt.com/}.

\bibitem[Taori et~al.(2023)Taori, Gulrajani, Zhang, Dubois, Li, Guestrin, Liang, and Hashimoto]{alpaca}
Rohan Taori, Ishaan Gulrajani, Tianyi Zhang, Yann Dubois, Xuechen Li, Carlos Guestrin, Percy Liang, and Tatsunori~B. Hashimoto.
\newblock Stanford alpaca: An instruction-following llama model.
\newblock \url{https://github.com/tatsu-lab/stanford_alpaca}, 2023.

\bibitem[Teknium(2023)]{OpenHermes_2_5}
Teknium.
\newblock Openhermes 2.5: An open dataset of synthetic data for generalist llm assistants, 2023.
\newblock URL \url{https://huggingface.co/datasets/teknium/OpenHermes-2.5}.

\bibitem[Touvron et~al.(2023)Touvron, Lavril, Izacard, Martinet, Lachaux, Lacroix, Rozi{\`e}re, Goyal, Hambro, Azhar, et~al.]{touvron2023llama}
Hugo Touvron, Thibaut Lavril, Gautier Izacard, Xavier Martinet, Marie-Anne Lachaux, Timoth{\'e}e Lacroix, Baptiste Rozi{\`e}re, Naman Goyal, Eric Hambro, Faisal Azhar, et~al.
\newblock Llama: Open and efficient foundation language models.
\newblock \emph{arXiv preprint arXiv:2302.13971}, 2023.

\bibitem[Tunstall et~al.(2023)Tunstall, Beeching, Lambert, Rajani, Huang, Rasul, Rush, and Wolf]{alignment_handbook2023}
Lewis Tunstall, Edward Beeching, Nathan Lambert, Nazneen Rajani, Shengyi Huang, Kashif Rasul, Alexander~M. Rush, and Thomas Wolf.
\newblock The alignment handbook.
\newblock \url{https://github.com/huggingface/alignment-handbook}, 2023.

\bibitem[Wang et~al.(2020)Wang, Wei, Dong, Bao, Yang, and Zhou]{wang2020minilm}
Wenhui Wang, Furu Wei, Li~Dong, Hangbo Bao, Nan Yang, and Ming Zhou.
\newblock Minilm: Deep self-attention distillation for task-agnostic compression of pre-trained transformers, 2020.

\bibitem[Wang et~al.(2022)Wang, Kordi, Mishra, Liu, Smith, Khashabi, and Hajishirzi]{wang2022selfinstruct}
Yizhong Wang, Yeganeh Kordi, Swaroop Mishra, Alisa Liu, Noah~A Smith, Daniel Khashabi, and Hannaneh Hajishirzi.
\newblock Self-instruct: Aligning language model with self generated instructions.
\newblock \emph{arXiv preprint arXiv:2212.10560}, 2022.

\bibitem[Wang et~al.(2023)Wang, Ivison, Dasigi, Hessel, Khot, Chandu, Wadden, MacMillan, Smith, Beltagy, et~al.]{wang2023far}
Yizhong Wang, Hamish Ivison, Pradeep Dasigi, Jack Hessel, Tushar Khot, Khyathi Chandu, David Wadden, Kelsey MacMillan, Noah~A Smith, Iz~Beltagy, et~al.
\newblock How far can camels go? exploring the state of instruction tuning on open resources.
\newblock \emph{Advances in Neural Information Processing Systems}, 36:\penalty0 74764--74786, 2023.

\bibitem[Wei et~al.(2021)Wei, Bosma, Zhao, Guu, Yu, Lester, Du, Dai, and Le]{wei2021finetuned-FLAN}
Jason Wei, Maarten Bosma, Vincent Zhao, Kelvin Guu, Adams~Wei Yu, Brian Lester, Nan Du, Andrew~M Dai, and Quoc~V Le.
\newblock Finetuned language models are zero-shot learners.
\newblock In \emph{International Conference on Learning Representations}, 2021.

\bibitem[Xu et~al.(2023)Xu, Sun, Zheng, Geng, Zhao, Feng, Tao, and Jiang]{xu2023wizardlm}
Can Xu, Qingfeng Sun, Kai Zheng, Xiubo Geng, Pu~Zhao, Jiazhan Feng, Chongyang Tao, and Daxin Jiang.
\newblock Wizardlm: Empowering large language models to follow complex instructions.
\newblock \emph{arXiv preprint arXiv:2304.12244}, 2023.

\bibitem[Xu et~al.(2022)Xu, Chen, Du, Shao, Yanggang, Li, and Yang]{xu2022zeroprompt}
Hanwei Xu, Yujun Chen, Yulun Du, Nan Shao, Wang Yanggang, Haiyu Li, and Zhilin Yang.
\newblock Zeroprompt: Scaling prompt-based pretraining to 1,000 tasks improves zero-shot generalization.
\newblock In \emph{Findings of the Association for Computational Linguistics: EMNLP 2022}, pages 4235--4252, 2022.

\bibitem[Xu et~al.(2024)Xu, Jiang, Niu, Deng, Poovendran, Choi, and Lin]{xu2024magpie}
Zhangchen Xu, Fengqing Jiang, Luyao Niu, Yuntian Deng, Radha Poovendran, Yejin Choi, and Bill~Yuchen Lin.
\newblock Magpie: Alignment data synthesis from scratch by prompting aligned llms with nothing, 2024.

\bibitem[Yue et~al.(2023)Yue, Qu, Zhang, Fu, Huang, Sun, Su, and Chen]{yue2023mammoth}
Xiang Yue, Xingwei Qu, Ge~Zhang, Yao Fu, Wenhao Huang, Huan Sun, Yu~Su, and Wenhu Chen.
\newblock Mammoth: Building math generalist models through hybrid instruction tuning.
\newblock In \emph{The Twelfth International Conference on Learning Representations}, 2023.

\bibitem[Zellers et~al.(2019)Zellers, Holtzman, Bisk, Farhadi, and Choi]{zellers2019hellaswag}
Rowan Zellers, Ari Holtzman, Yonatan Bisk, Ali Farhadi, and Yejin Choi.
\newblock Hellaswag: Can a machine really finish your sentence?
\newblock In \emph{Proceedings of the 57th Annual Meeting of the Association for Computational Linguistics}, 2019.

\bibitem[Zhang et~al.(2024)Zhang, Schwarzschild, Carlini, Kolter, and Ippolito]{zhang2024forcing}
Yiming Zhang, Avi Schwarzschild, Nicholas Carlini, Zico Kolter, and Daphne Ippolito.
\newblock Forcing diffuse distributions out of language models.
\newblock \emph{arXiv preprint arXiv:2404.10859}, 2024.

\end{thebibliography}

\appendix
\clearpage
\section{Appendix}

\subsection{Finetuning Hyperparameters}\label{sec:ft-hparams}

We utilize the Hugging Face Alignment Handbook codebase \cite{alignment_handbook2023} for our finetuning runs and the same set of standard hyperparameters for each dataset. During finetuning, we employ AdamW \citep{loshchilov2017decoupled} without weight decay. We warm up for the first $10\%$ of the total steps, and after the warm-up period, we use the learning rate to $1\text{e-}6$ and utilize a cosine annealing scheduler, reducing it to $0$. Our setup includes a batch size of $8$ per device, an accumulation step of $2$ and a number epoch of $1$ with a sequence length of $2048$. All experiments are conducted on a single node equipped with $8$ NVIDIA H100 cards, resulting in a global batch size of $128$ per gradient update. Additionally, the percent of each split in the rebalanced GenQA can be seen in \Cref{table:ration}.


\begin{table}[!h]
\setlength{\tabcolsep}{4.0pt}
\centering
\caption{Percent of each split in the rebalanced GenQA. The rebalanced dataset comprises 6,470,575 questions, for 10,906,517 turns across all conversations.  }\label{tab:rebalance-ratios}
\label{table:ration}
\begin{tabular}{ccccccccc}
\toprule
Code & General & Task & Writing & Dialog & Math & Multiple choice & Academic & MMLU \\
\midrule
8\% & 5\% & 16\% & 14\% & 13\% & 8\% & 6\% & 15\% & 15\% \\
\bottomrule
\end{tabular}
\end{table}

\subsection{Details and Example Questions of Each Split} 

We briefly describe the content of each split and how it was created.  We also present one selected question from each split to give the reader a sense of the diversity and quality of questions present in GenQA.  

Each split was created for forming a generator-type ``meta prompt,'' which was then fed to the Gemini 1.0 Pro LLM many times to produce different outputs.  Each entry in the dataset was produced by a separate API call (for multi-turn conversations, all turns were created at once).  Some splits used prompts that are conditioned on a randomly chosen topic.  The lists from which these topics were selected are show in Appendix \ref{sec:topics}. 

\label{sec:represent}
\textbf{Academic}  This split was created by asking Gemini 1.0 Pro to create a question and answer from scratch.  Generator prompting was used, both with and without conditioning on a pre-defined list of topics that Gemini says it can write about.
An example question from this split is as follows. 

\begin{tcolorbox}[breakable,boxrule=0pt]
{
{\em USER:}  What are the unique characteristics and challenges associated with investigating and prosecuting domestic violence cases?
\\\\
{\em ASSISTANT:}
Domestic violence cases present several complexities and challenges for investigators and prosecutors:
\begin{itemize}
\item \textbf{Intimate Nature of the Relationship:} The close and personal nature of domestic violence relationships makes it difficult for victims to report and seek help. They may feel ashamed, embarrassed, or fearful of retaliation.
\item \textbf{Lack of Physical Evidence:} In many cases, domestic violence does not leave visible physical injuries or clear evidence of a crime. This can make it difficult to prove the offense and hold the offender accountable.
\item \textbf{Power Imbalance:} Domestic violence relationships are often characterized by power imbalances, with one person having more physical, emotional, or financial control than the other. This power dynamic makes it difficult for victims to escape or resist abuse.
\item \textbf{Cycle of Violence:} Domestic violence often follows a cycle of violence, where episodes of abuse are followed by periods of calm or reconciliation. This cycle can make it difficult for victims to break free from the relationship and obtain safety.
\item \textbf{High Risk of Re-victimization:} Victims of domestic violence are at a significantly increased risk of being re-victimized. This risk is exacerbated by the fact that offenders often have multiple contacts with the criminal justice system, which provides opportunities for them to continue abusing their victims.
\item \textbf{Challenges in Prosecution:} Domestic violence cases can be difficult to prosecute because of the lack of physical evidence, the reluctance of victims to testify, and the legal defenses often raised by offenders, such as provocation or self-defense.
\end{itemize}

}
\end{tcolorbox}

The question above was generated using the following meta-prompt.  Note that the question is conditioned on the topic ``Crime and deviance,'' which was randomly chosen at generation time.

    \texttt{List 40 subtopics in the domain of Crime and deviance.   State subtopic 14.  Then write a question that is not about subtopic 14, but can only be answered with expertise in subtopic 14, and then write the answer. Both the question and answer should be long. The name of the subtopic should not appear in the question.  Begin your questions with "Question:" and your answer with "Answer:".}

\textbf{MMLU}  This split was created by asking Gemini 1.0 Pro to create a question and answer from scratch.  Generator prompting was used, but the model was instructed to focus on topics that appear in MMLU. Note that these questions, while being on MMLU topics, are not necessarily formatted in the style or MMLU, and are not meant to be representative of MMLU questions.

\begin{tcolorbox} [breakable,boxrule=0pt]
{\em USER:} Explain the complex geopolitical factors that led to the outbreak of the Yugoslav Wars. Consider the rise of nationalism, ethnic tensions, and the role of foreign intervention.
\\ \\
    {\em ASSISTANT:} The Yugoslav Wars erupted due to a confluence of geopolitical factors. Following the death of Tito, Yugoslavia's charismatic leader, the country faced a power vacuum and a resurgence of nationalism. Ethnic tensions between Serbs, Croats, Bosniaks, and Albanians escalated, fueled by historical grievances and territorial disputes. External powers, including the United States and Russia, played a significant role in arming and supporting different factions. The European Union, struggling to maintain unity, initially failed to intervene effectively, allowing the conflict to spiral into a bloody civil war.
\end{tcolorbox}

    This question was generated from the following meta-prompt.

    \texttt{List 40 subtopics in the domain of High School European History.   State subtopic 25.  Then write a question that is not about subtopic 25, but can only be answered with expertise in subtopic 25, and then write the answer. Both the question and answer should be long. The name of the subtopic should not appear in the question.  Begin your questions with "Question:" and your answer with "Answer:".  Be creative and don't ask the first thing you think of.}

\textbf{Multiple Choice} This split has similar content to Academic and MMLU, but questions are presented in multiple choice format.  Answers contain not only the letter/number corresponding to the answer, but also a textual explanation for the choice.

\begin{tcolorbox}[breakable,boxrule=0pt]
{\em USER:} Which of the following is NOT a characteristic of an evolutionary arms race?
    
    A. Reciprocal adaptations between two populations
    
    B. Escalation of the intensity of adaptations
    
    C. Stabilizing selection
    
    D. Negative frequency-dependent selection
    \\\\
    {\em ASSISTANT:} C
    
    \textbf{Explanation:}
    
    In an evolutionary arms race, both populations are constantly adapting to each other, leading to a reciprocal escalation of adaptations. Stabilizing selection, however, favors the maintenance of a constant trait value and would not be present in an evolutionary arms race.
 \end{tcolorbox}

    The question was created using the following meta-prompt:
    
\texttt{List 40 subtopics in the domain of Evolution.   Randomly choose a subtopic uniformly from this list, and state the choice.  Then write a long complex multiple-choice question that is not about the subtopic, but can only be answered with expertise in the subtopic. The question should end with a list of choices.  Then write the answer, followed by an explanation of your choice. The name of the subtopic should not appear in the question.  Begin your questions with "Question:" and your answer with "Answer:".  Don't ask the first thing you think of.}

\textbf{Writing}  These questions ask the assistant to write a document.  Questions are conditioned on a document type from a list of possible document types that was pre-written by Gemini.

\begin{tcolorbox}[breakable,boxrule=0pt]

{\em USER:}  Create a shipping label for a package containing a laptop being sent from New York City to Los Angeles. The package weighs 5 pounds and is valued at \$2,000. 
\\\\
{\em ASSISTANT:}

\textbf{Shipping Label}

\textbf{From:}

John Doe

123 Main Street

New York City, NY 10001

\textbf{To:}

Jane Smith

456 Elm Street

Los Angeles, CA 90001

\textbf{Package Details:}
\begin{itemize}
    \item  Laptop
    \item  Weight: 5 pounds
    \item Value: \$2,000
\end{itemize}

\textbf{Shipping Method:}

    USPS Priority Mail
\\\\
{\em USER:} Imagine that the package is containing a gift for Jane's birthday. Write a short story about the journey of the package from New York City to Los Angeles, including the adventures it encounters along the way.
\\\\
{\em ASSISTANT:}     \textbf{The Package's Journey}

Once upon a time, in the bustling metropolis of New York City, a small package containing a special gift was carefully packed and labeled. Its destination was the sunny shores of Los Angeles, where Jane eagerly awaited its arrival.

Little did the package know that its journey would be anything but ordinary. As it was loaded onto the delivery truck, it couldn't help but feel a twinge of excitement. The adventure had begun!

The truck hummed along the busy streets, passing towering skyscrapers and bustling crowds. But its journey took a detour when it was caught in a torrential downpour. The rain lashed against the metal exterior of the truck, and the package swayed precariously inside.

Undeterred, the package continued its journey. It traveled through rolling hills and vast plains, witnessing the changing landscapes of the country. It shared a ride with a chatty group of parcels, each with their own story to tell.

As the package approached Los Angeles, it found itself in a traffic jam. The sun beat down relentlessly, but the package remained inside its protective wrapping, shielded from the heat.

Finally, after a long and eventful journey, the package arrived at Jane's doorstep. She was overjoyed to receive it and couldn't wait to open it.

Inside the package, Jane found a beautiful necklace adorned with twinkling diamonds. It was the perfect gift, and it held a special meaning for her. And so, the package's journey came to an end, leaving behind a trail of memories and a cherished treasure for Jane.
\end{tcolorbox}

The question was created using the following meta-prompt.
    
    \texttt{List 15 writing prompts that ask for a specific document of the following type: shipping label. Begin your list with the header "Topics:", and number the topics 1-15. Then print "Writing Prompt:", followed by a complete writing prompt about topic 8. Then write a passage on topic 8. Begin your passage with "Passage:".\u00a0Then, write 20 different types of questions you could ask about this passage.  Begin this list with "Question Types:". Number the types in this list 1-20.  Then state type 5.  Then write an instruction of type 5 about the passage. Do not include the name of type 5 in your instruction.  Begin your instruction with "Instruction:". Then write a response to your instruction.  Begin your response with "Response:". Your passage and response should both be long.}

\textbf{Task}
This split contains a range of text-based tasks, most of which do not involve compositional writing. Tasks are generated on-the-fly by generator prompting, and are not drawn from a fixed list. Example tasks include summarizing, fill in the blank, trip planning, text categorization, text analysis, editing, and more.  Many of the examples in this split are multi-turn. 

\begin{tcolorbox}[breakable,boxrule=0pt]

    {\em USER:} Generate a natural language description of the following JSON data:
    
    \begin{verbatim}
{
  "name": "John Doe",
  "age": 25,
  "occupation": "Software Engineer",
  "hobbies": ["coding", "reading", "traveling"],
  "education": [
    {
      "degree": "Bachelor of Science in Computer Science",
      "university": "Stanford University",
      "graduation_year": 2019
    }
  ]
}
\end{verbatim}
    {\em ASSISTANT:} John Doe is a 25-year-old software engineer. He holds a Bachelor of Science in Computer Science from Stanford University, graduating in 2019. John enjoys coding, reading, and traveling.
    \\\\
    {\em USER:} Generate the natural language description in a different format, such as a bullet list or table.
    \\\\
    {\em ASSISTANT:}  
    \begin{itemize}
    \item Name: John Doe
    \item Age: 25
    \item Occupation: Software Engineer
    \item Hobbies:
    \begin{itemize}
        \item Coding
        \item Reading
        \item Traveling
    \end{itemize}
    \item Education:
    \begin{itemize}
        \item Bachelor of Science in Computer Science, Stanford University (2019)
    \end{itemize}
\end{itemize}
\end{tcolorbox}

This question was produced by the following meta-prompt.

\texttt{List 30 random types of text-based tasks.  Then choose a random task from this list, and state your choice and its number.  Then write a random instruction to perform a random task of that type.  If the task involves a passage of text, include the passage in the instruction.  The instruction should be self-contained, and should not contain any text from the task that is required to perform the instruction, and should not involve an image.  Then write a response to the instruction. Both the question and answer should be long.  Begin your instruction with "Instruction:" and your response with "Response:".  Be creative.}

\textbf{Code}
This split contains questions about writing code in various languages, followed by answers that typically contain a block of code.

\begin{tcolorbox}[breakable,boxrule=0pt]
    {\em USER:}  How can I write a Java program that allows a user to input a list of numbers and calculates the average of the numbers?.
    \\\\
    {\em ASSISTANT:} Here's a way to write a Java program that calculates the average of a list of numbers inputted by the user:
    
 \begin{lstlisting}
import java.util.Scanner;

public class AverageCalculator {    
    public static void main(String[] args) {
        Scanner input = new Scanner(System.in);
        
        // Create a list to store the numbers        
        List<Integer> numbers = new ArrayList<>();
        
        // Prompt the user to enter numbers
        System.out.println(
            "Enter numbers (type 'q' to quit):"
        );
        
        // Read numbers from the user until they enter 'q'
        while (input.hasNextInt()) {
            numbers.add(input.nextInt());
        }
        // Calculate the average of the numbers
        double average = calculateAverage(numbers);
        // Print the average
        System.out.println("The average of the numbers is: " 
            + average);
    }
    private static double calculateAverage(
                                List<Integer> numbers) {
        double sum = 0;
        for (int number : numbers) {
            sum += number;
        }        
        return sum / numbers.size();
    }
}
       \end{lstlisting}
In this program, we use a `Scanner` to read numbers from the user and store them in a list. We use a `while` loop to continue reading numbers until the user enters 'q'. Then, we call the `calculateAverage()` method to calculate the average of the numbers in the list. Finally, we print the average.
    
\end{tcolorbox}
The  question was generated using the following meta-prompt.

\texttt{Write a random question about how to do something complex using Java.  Then write the answer to the question including examples.  Begin your question with "Question:" and your answer with "Answer:".  Be creative.  }

\textbf{Math}
This split contains multi-turn conversations about Math questions and answers, with difficulty ranging from elementary to graduate level. Each sample is labelled with its difficulty level.

\begin{tcolorbox}[breakable,boxrule=0pt]
 
 {\em USER:}  Determine if the following set of vectors forms a basis for $R^3$:
 
 $$v1 = [1, 2, -1]$$
 $$ v2 = [3, 1, 2]$$
 $$ v3 = [2, 3, 1]$$

{\em ASSISTANT:} Step 1: Check Linear Independence 

To determine if the vectors form a basis, we must check linear independence. We set up the equation:
$$a[1, 2, -1] + b[3, 1, 2] + c[2, 3, 1] = 0 $$
and solve for the coefficients $a$, $b$, and $c$.

Solving this system of equations gives $a = b = c = 0$, indicating that the vectors are linearly independent.

Step 2: Check Spanning 

We must also check if the vectors span $R^3$. This means that every vector in $R^3$ can be written as a linear combination of $v1$, $v2$, and $v3$. 

One way to check this is to set up the augmented matrix:

\[
\begin{bmatrix}
1 & 3 & 2 & \vert & 1 \\
2 & 1 & 3 & \vert & 0 \\
-1 & 2 & 1 & \vert & 0
\end{bmatrix}
\]

Row reducing this matrix gives the identity matrix, indicating that every row vector in the original matrix can be written as a linear combination of the three column vectors.

Conclusion: 

Since the vectors are linearly independent and span $R^3$, they form a basis for $R^3$.

 {\em USER:}  Find a basis for the null space of the matrix:
\[
A = \begin{bmatrix}
1 & 2 & 3 \\
4 & 5 & 6 \\
7 & 8 & 9
\end{bmatrix}
\]

{\em ASSISTANT:}  Step 1: Find the Row Echelon Form of A 

Solving for the row echelon form of A gives:
\[
A = \begin{bmatrix}
1 & 0 & -1 \\
0 & 1 & 2 \\
0 & 0 & 0
\end{bmatrix}
\]

Step 2: Identify the Free Variables 

The last column of the row echelon form has a free variable, which we label as z.

Step 3: Express Basic Variables in Terms of Free Variables 

From the row echelon form, we have:
$$x = -z$$
$$y = -2z$$

Step 4: Write the Null Space Basis 

The null space basis is given by the solutions to the homogeneous system $Ax = 0$. Using the above expressions, we have:

$$\text{Null Space Basis} = \{(z, -2z, z) | z \in R\}$$
\end{tcolorbox}

This question was made using the following meta-prompt.

\texttt{List 20 topics of math problems.  Then, choose a topic at random, and print the name of this topic.  Then list 20 sub-topics.  Then, choose one of these at random.  Then write a complex question that does not include the name of the sub-topic, but requires expertise in the sub-topic to answer.  Then write the detailed answer and show your work.  Begin your question with "Question:" and your answer with "Answer:".  Then, write a follow-up question.  Label your follow up question "Question2:".  Then write the detailed follow-up answer. Show your work.  Label the follow-up answer "Answer2:".   Then, write another follow-up question, labelled "Question3:".  Then write the detailed follow-up answer,  labelled  "Answer3:".  Show your work.   Then, print "Difficulty:", followed by a classification of the difficulty of the math problems as "elementary", "high school", "college", or "graduate".}

\textbf{Dialog}
These casual multi-turn conversations between a chatbot and a human cover a range of topics.

\begin{tcolorbox}[breakable,boxrule=0pt]

{\em USER:}  So, let's talk about the physics behind skateboarding tricks. I've always been curious about how skateboarders are able to perform such incredible aerial maneuvers. 
\\\\
{\em ASSISTANT:} Skateboarding tricks are indeed fascinating from a physics standpoint. One key element is momentum. When a skateboarder pushes forward, their body and the board gain momentum. This momentum can be used to launch them into the air. 
\\\\
{\em USER:}  That makes sense. But how do they control their rotation in the air? 
\\\\
{\em ASSISTANT:} Rotation is achieved through angular momentum. When the skateboarder shifts their weight or moves their body, they create a torque that causes the board to rotate. They can control the direction and speed of the rotation by adjusting their movements. 
\\\\
{\em USER:} Amazing! What about the landings? How do they avoid crashing? 
\\\\
{\em ASSISTANT:}  Landing safely requires precise timing and balance. Skateboarders use the principles of gravity and friction. As they approach the landing, they need to adjust their body position to reduce their speed and distribute the impact. Friction between the board and the ground helps them slow down and regain stability. 
\\\\
{\em USER:} This is really interesting. What are some of the most challenging tricks to perform? 
\\\\
{\em ASSISTANT:}  Some of the most challenging tricks involve multiple rotations, flips, and inversions. The "900" is a trick where the skateboarder rotates their body and board three full times in the air. The "McTwist" is a trick where the skateboarder rotates their body 540 degrees while flipping the board under their feet. These tricks require incredible skill, precision, and athleticism.  
\end{tcolorbox}

This question came from the following meta-prompt.

\texttt{Write a long multi-turn dialog between a user and an AI assistant.  Start by printing a numbered list of 30 random concrete topics of conversations that could take place between the user and assistant. None of the topics should be about current events, local business recommendations, or the abstract issues like the meaning the life.  Then print "Selected topic:" followed by the name of a topic chosen randomly from the list.  Then, write the long multi-turn dialog.  Begin each user statement with "User:" and each assistant statement with "Assistant:".  Each assistant response should be long. There should be 4 or more turns between the user and assistant.}

\textbf{General}
This experimental dataset was created by handing GPT-3.5 an empty string, to which it typically responds with an answer to some (unknown) instruction.  After generating the answer in this way, the question corresponding to this answer was written by Gemini.  

\begin{tcolorbox}[breakable,boxrule=0pt]  

{\em USER:}  Will the Xbox Series X be backwards compatible with games for the Xbox One and Xbox 360? 
\\\\
{\em ASSISTANT:}  No, the Xbox Series X will not be backwards compatible with Xbox One or Xbox 360 games. However, Microsoft has stated that they are working on improving backwards compatibility for older games on the Xbox Series X. 
\end{tcolorbox}

\subsection{Diversity Analysis}\label{diversity-app}
\begin{figure}[!h]
    \centering
    \includegraphics[width=0.49\linewidth]{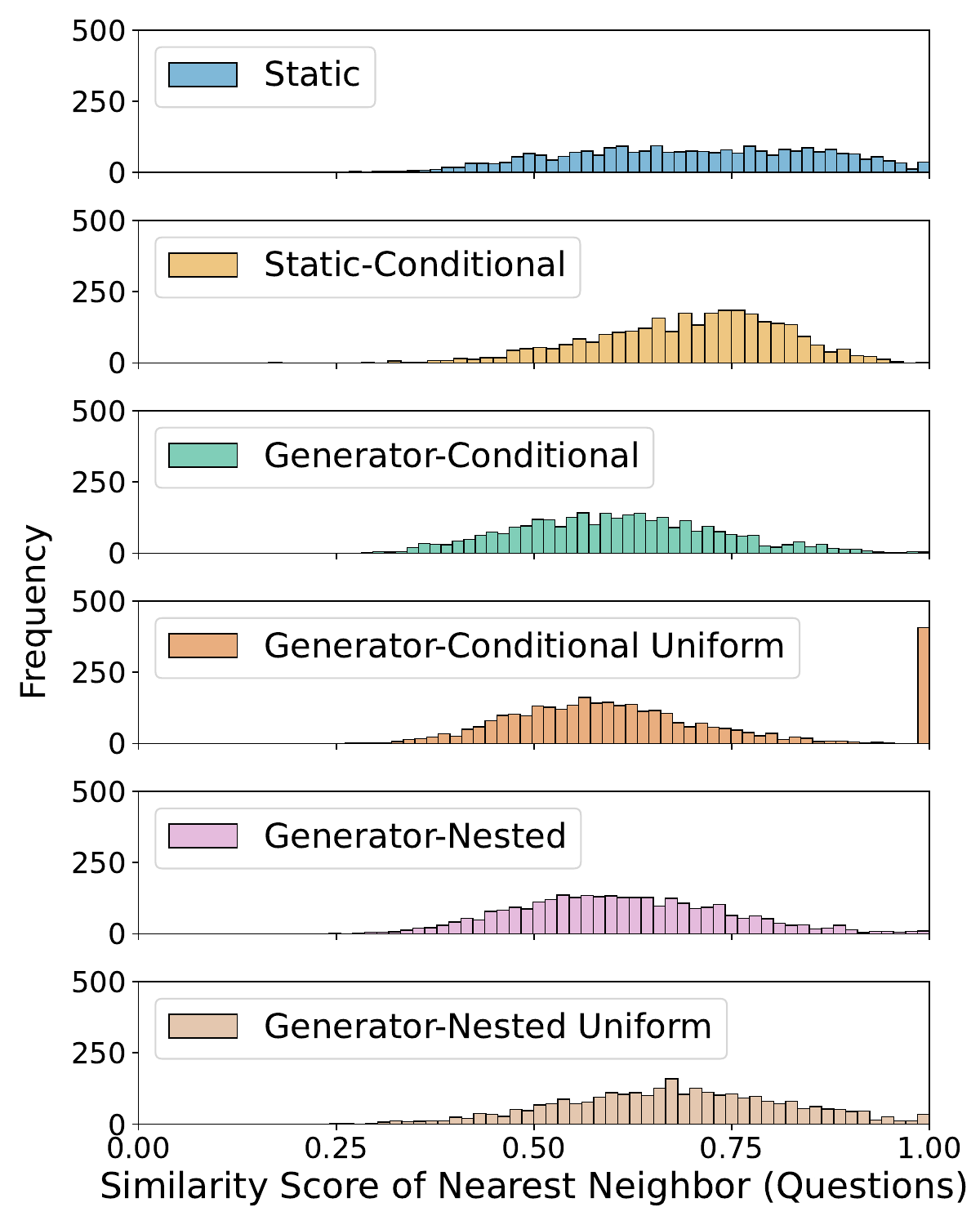}
    \includegraphics[width=0.49\linewidth]{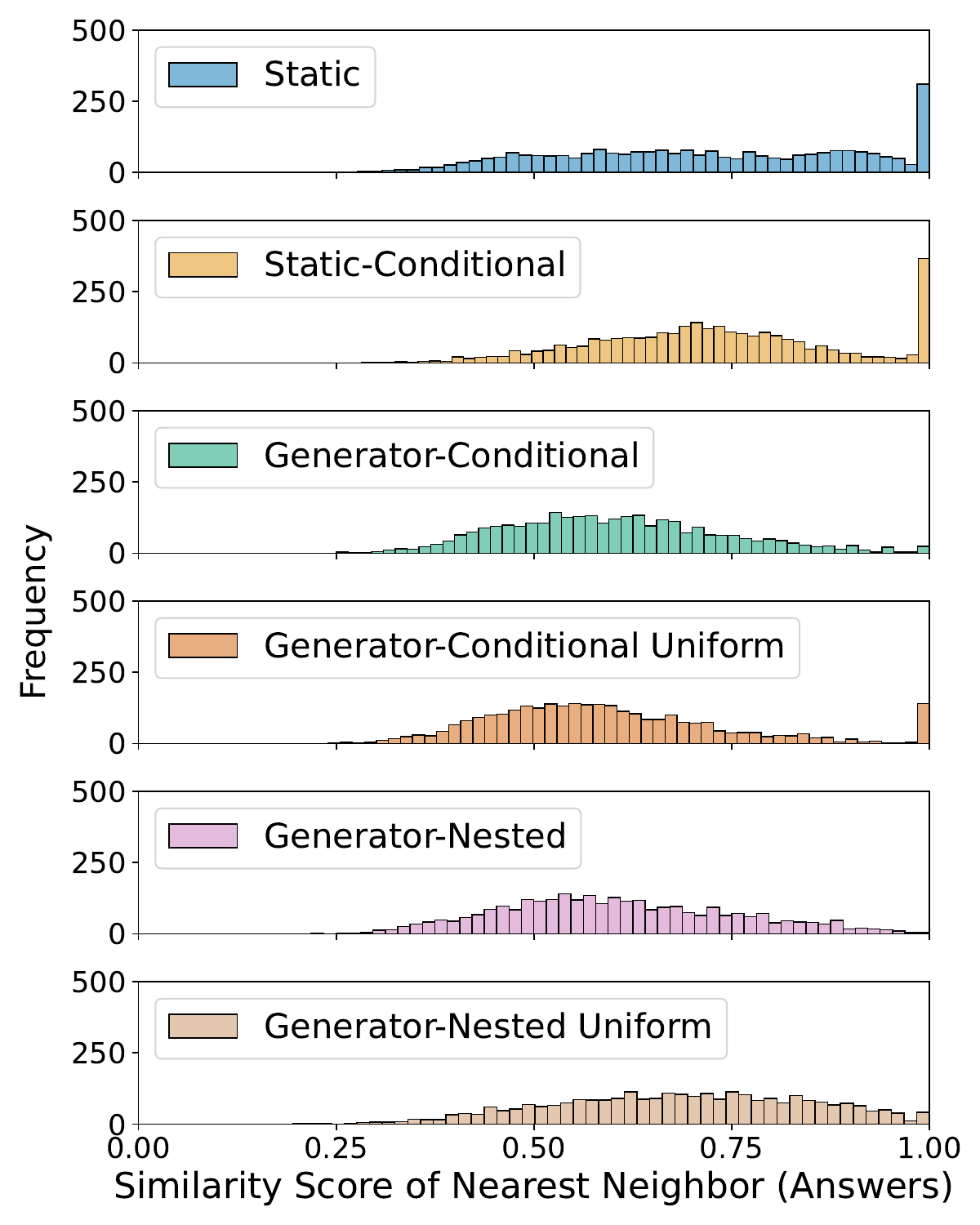}
    \caption{Comparison of nearest neighbor ($k=1$). Similarity of questions and answers in Academic split.}
    \label{fig:academic-full}
\end{figure}

To evaluate which prompting strategy produces the most diverse pairs of questions and answers, we propose to examine the diversity of our dataset by analyzing the similarities among nearest-neighbor of embeddings derived from questions and answers.
As discussed in \Cref{sec:generator-science}, all prompting strategies are evaluated in \Cref{fig:academic-full}. To our surprise, giving the LLM the freedom to sample the chosen index from the list of topics does not result in a more random generation, suggesting a potential bias in the model's selection process. 

We list the prompts used for our experiment below:

\textbf{Static: }\texttt{Write a random complex question and its long answer. Begin your question with "Question:" and your answer with "Answer:".}

\textbf{Static-Conditional:  }\texttt{Write a complex question from the domain of {topic}.  Then write the long answer.  Your question should not contain the words "{topic}". Begin your question with "Question:" and your answer with "Answer:"} 

\textbf{Generator-Conditional: }\texttt{List 40 subtopics in the domain of \{random\_topic\}.   State subtopic \{ind\}.  Then write a question that is not about subtopic \{ind\}, but can only be answered with expertise in subtopic \{ind\}, and then write the answer. Both the question and answer should be long. The name of the subtopic should not appear in the question.  Begin your questions with "Question:" and your answer with "Answer:".\{booster\}"}

\textbf{Generator-Conditional-Uniform: }\texttt{List 40 subtopics in the domain of \{random\_topic\}.   Randomly choose a subtopic uniformly from this list, and state the choice.  Then write a long complex question that is not about the subtopic, but can only be answered with expertise in the subtopic. Then write the answer, followed by an explanation of your choice. The name of the subtopic should not appear in the question.  Begin your questions with "Question:" and your answer with "Answer:".\{booster\}"}

\textbf{Generator-Nested: }\texttt{List \{n1\} topics that you can answer questions about. State topic \{ind1\}.  Then write \{n2\} subtopics about topic {ind1}.  Then state the subtopic \{ind2\}.  Then write a question that is not about subtopic {ind2}, but can only be answered with expertise in subtopic \{ind2\}. Then write the answer. Both the question and answer should be long. The name of the subtopic \{ind2\} should not appear in the question, and none of the words in subtopic \{ind2\} should be reused in the question. Begin your questions with "Question:" and your answer with "Answer:".\{booster\}"}

\textbf{Generator-Nested-Uniform: }\texttt{List {n1} topics that you can answer questions about. Choose a topic uniformly from this list, and state it.  Then write 60 subtopics about the chosen topic.  Then choose a subtopic uniformly from this list, and state it.  Then write a question that is not about the subtopic, but can only be answered with expertise in the subtopic. Then write the answer. Both the question and answer should be long. The name of the subtopic should not appear in the question, and none of the words in subtopic should be reused in the question. Begin your questions with "Question:" and your answer with "Answer:".\{booster\}"}

\begin{figure}[!h]
    \centering
    \includegraphics[width=0.49\linewidth]{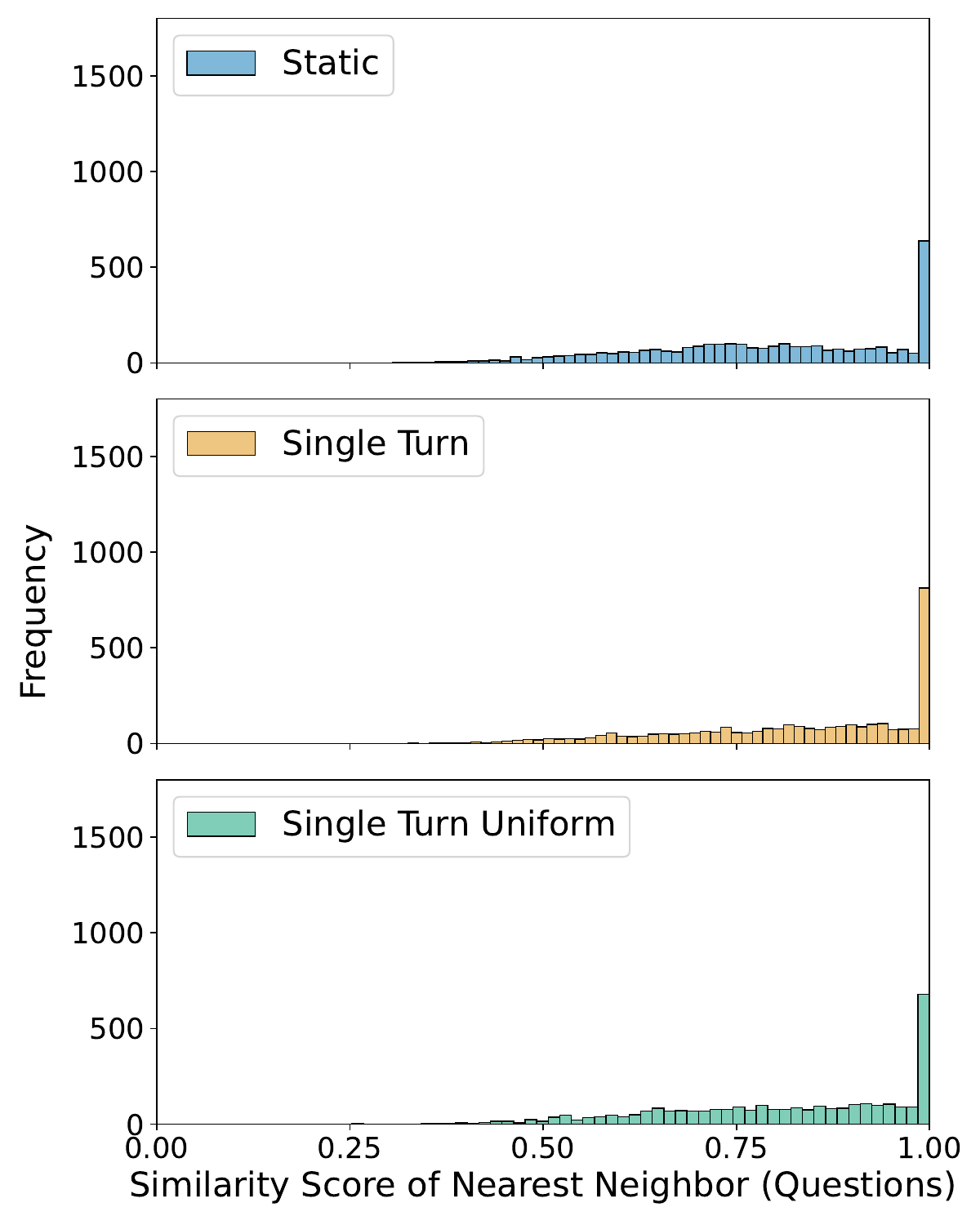}
    \includegraphics[width=0.49\linewidth]{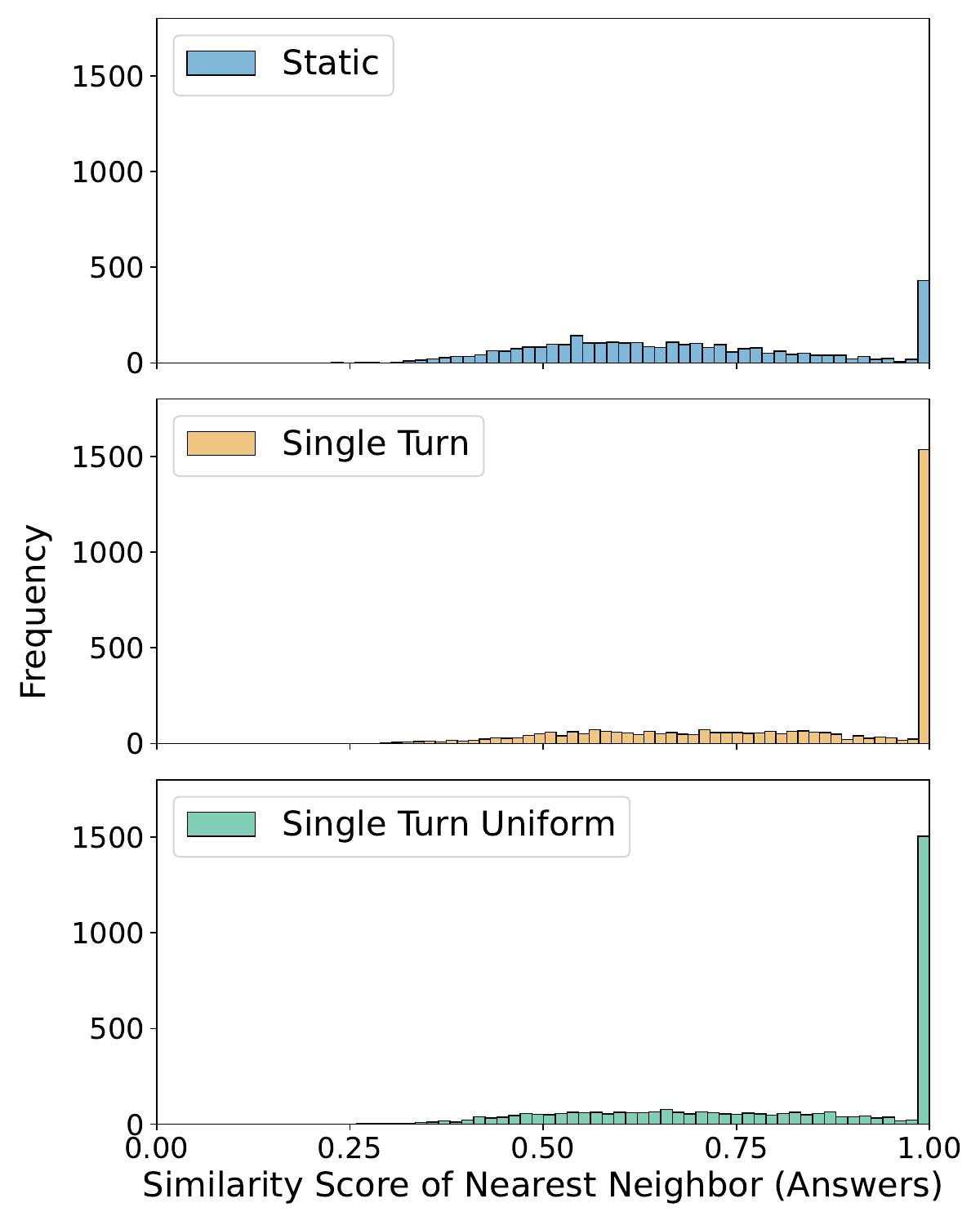}
    \caption{Comparison of nearest neighbor ($k=1$). Similarity of questions and answers in Task split.}
    \label{fig:taskk1}
\end{figure}

\Cref{fig:taskk1} shows the histograms of the similarity scores of the nearest neighbor of either the question or answer in the Task split. We list the prompts used for our experiment below. We generate 3000 examples of each prompt type and follow a similar set up as in \Cref{sec:generator-science}.

\textbf{Static: }\texttt{State a random type of text-based task.  Then write an instruction to perform a random task of that type.  The instruction should be self-contained, and should contain any text from the task that is required to perform the instruction. If the instructions refers to a passage of text, provide the passage in the instruction.  Then write a response to the instruction. Both the question and answer should be long. Begin your instruction with "Instruction:" and your response with "Response:".  Be creative.}

\textbf{Single Turn: }\texttt{List 30 random types of text-based tasks.  Then choose a random task from this list, and state your choice and its number.  Then write a random instruction to perform a random task of that type.  If the task involves a passage of text, include the passage in the instruction.  The instruction should be self-contained, and should not contain any text from the task that is required to perform the instruction, and should not involve an image.  Then write a response to the instruction. Both the question and answer should be long.  Begin your instruction with "Instruction:" and your response with "Response:".  Be creative.}

\textbf{Single Turn - Uniform: }\texttt{List 30 random types of text-based tasks.  Randomly choose a task uniformly from this list, and state your choice and its number.  Then write a random instruction to perform a random task of that type.  If the task involves a passage of text, include the passage in the instruction.  The instruction should be self-contained, and should not contain any text from the task that is required to perform the instruction, and should not involve an image.  Then write a response to the instruction. Both the question and answer should be long.  Begin your instruction with "Instruction:" and your response with "Response:".  Be creative.}

\subsection{Booster Effect} \label{sec:app-booster}
\begin{figure}[!h]
    \centering
    \includegraphics[trim={0.0cm 1.0cm 0.0cm 0.0cm},clip,width=0.8\linewidth]{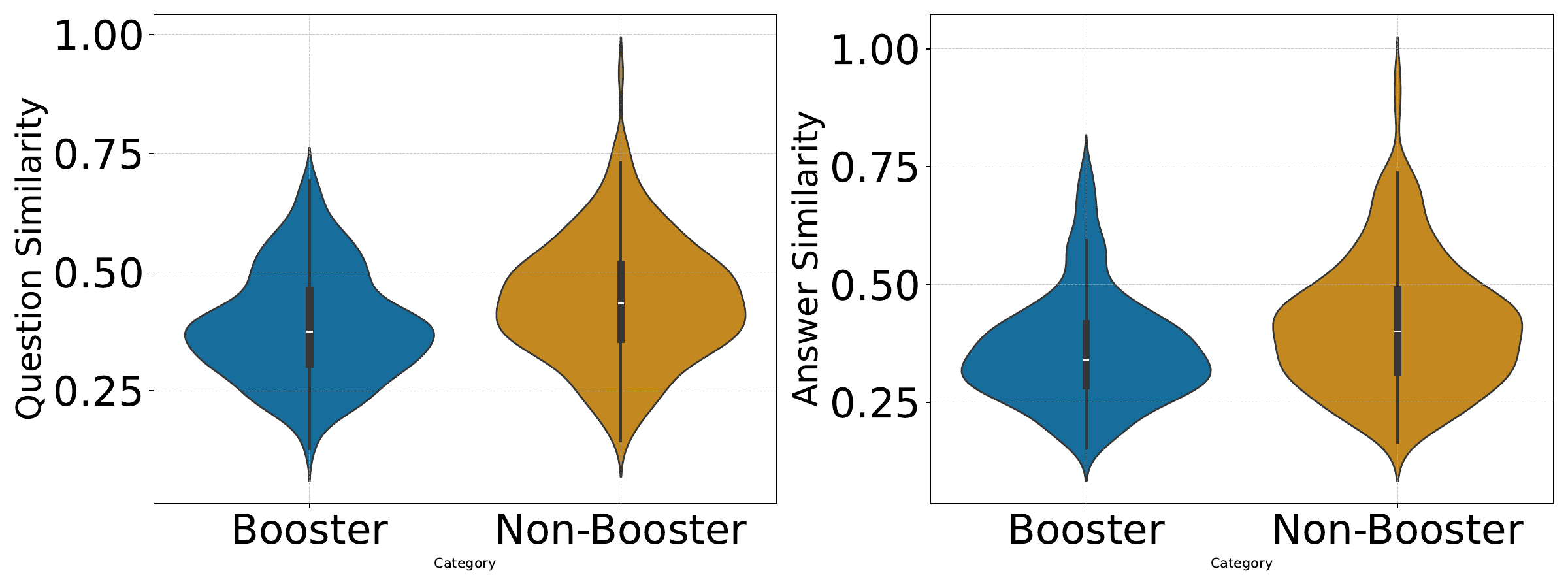}
    \includegraphics[trim={0.0cm 1.0cm 0.0cm 0.0cm},clip,width=0.8\linewidth]{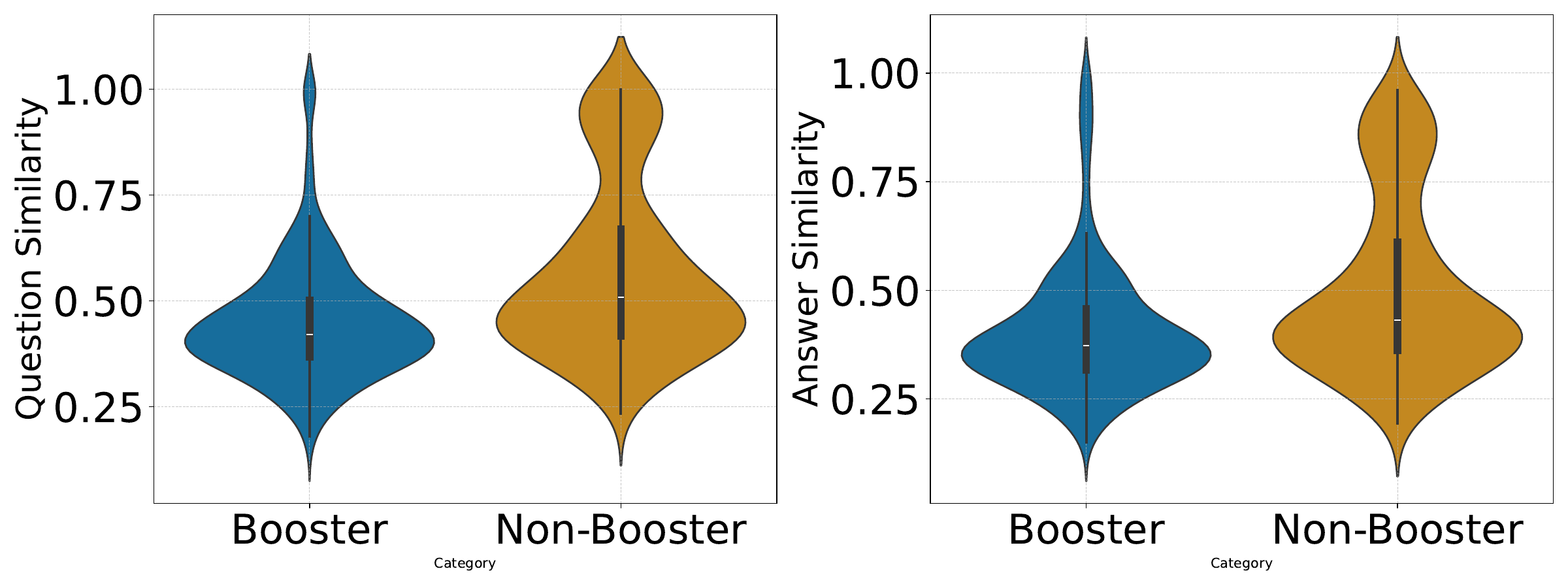}
    \includegraphics[trim={0.0cm 1.0cm 0.0cm 0.0cm},clip,width=0.8\linewidth]{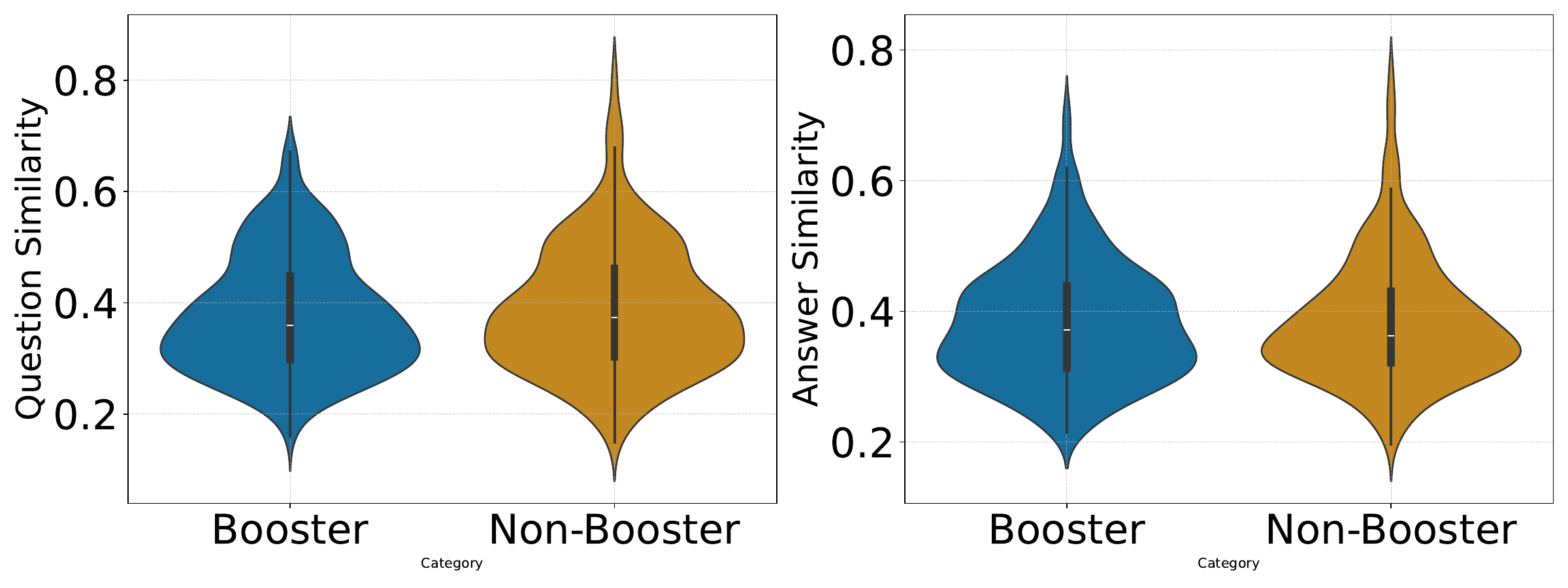}
    \caption{Booster Effect on the generated questions (right) and answers (left) in MMLU, Dialogue, and Multiple Choice splits. The distribution for the non-booster exhibits a longer tail approaching a similarity of $1$, indicating that the booster intervention contributes to an enhanced diversity in the generated outputs.}
    \label{fig:booster_effects}
\end{figure}

We evaluate the impact of using a booster in our prompting strategy by computing the similarity of the question or answer's nearest neighbor. As seen in \Cref{fig:booster_effects}, by using a booster in our prompt, we produce more unique pairs of questions and answers, with lower similarity scores observed across all splits.

\clearpage
\subsection{Token Analysis}\label{sec:token-analysis}
\begin{figure}[!h]
    \centering
    \includegraphics[width=0.49\linewidth]{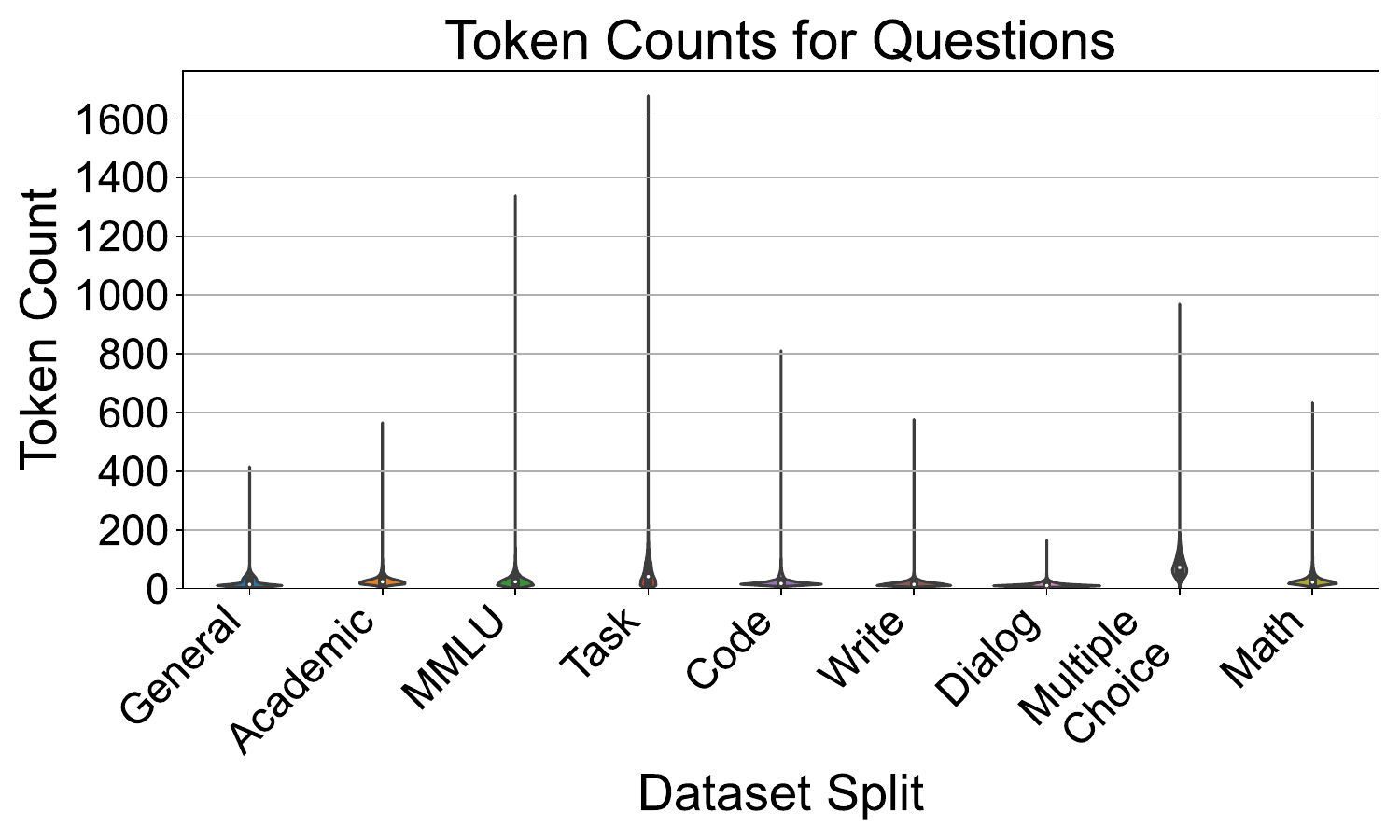}
    \includegraphics[width=0.49\linewidth]{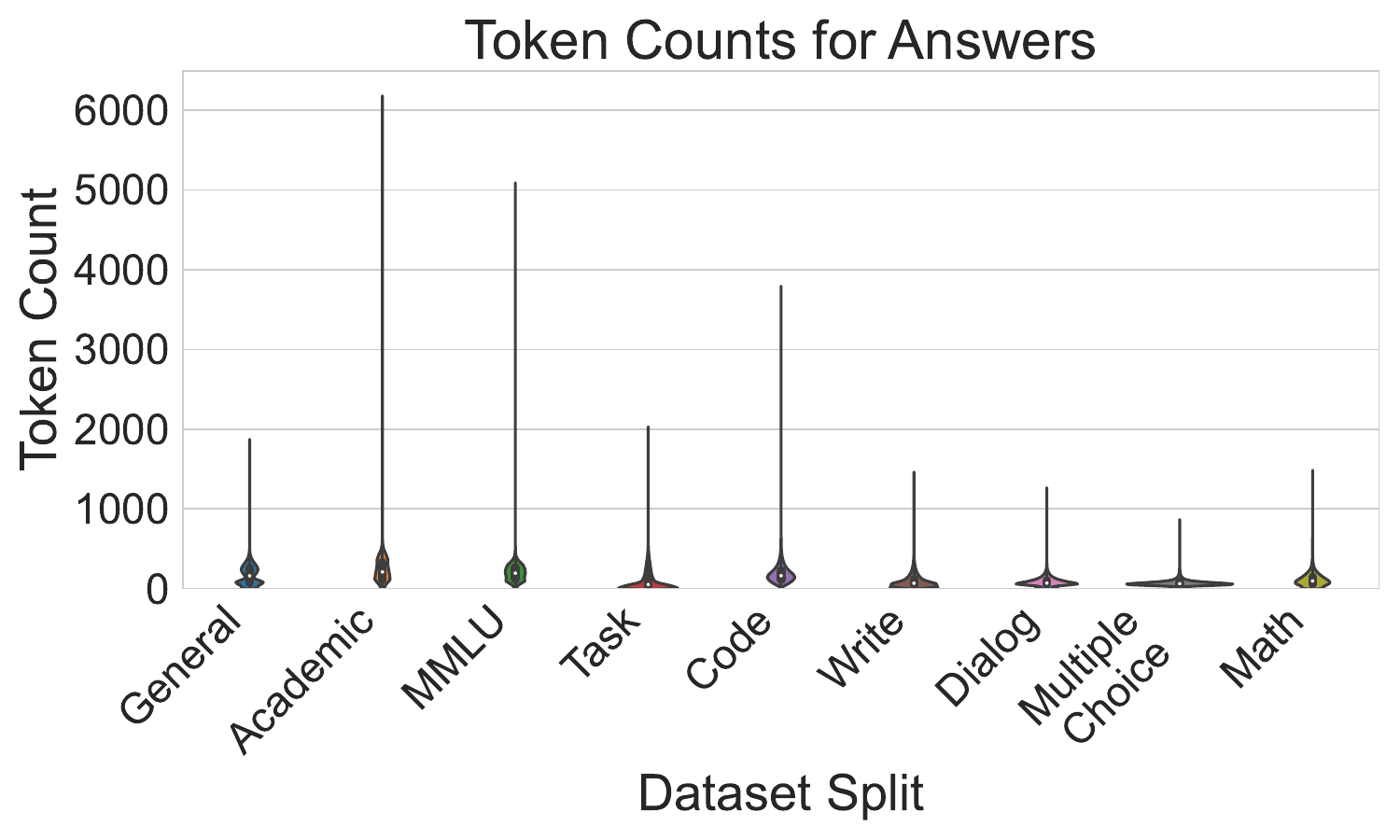}
    \caption{White-space token count for Questions (left) and Answers (right) for each of the nine spits.}
    \label{fig:white-space_token_count_full_ylim}
\end{figure}

\begin{figure}[!h]
    \centering
    \includegraphics[width=0.65\linewidth]{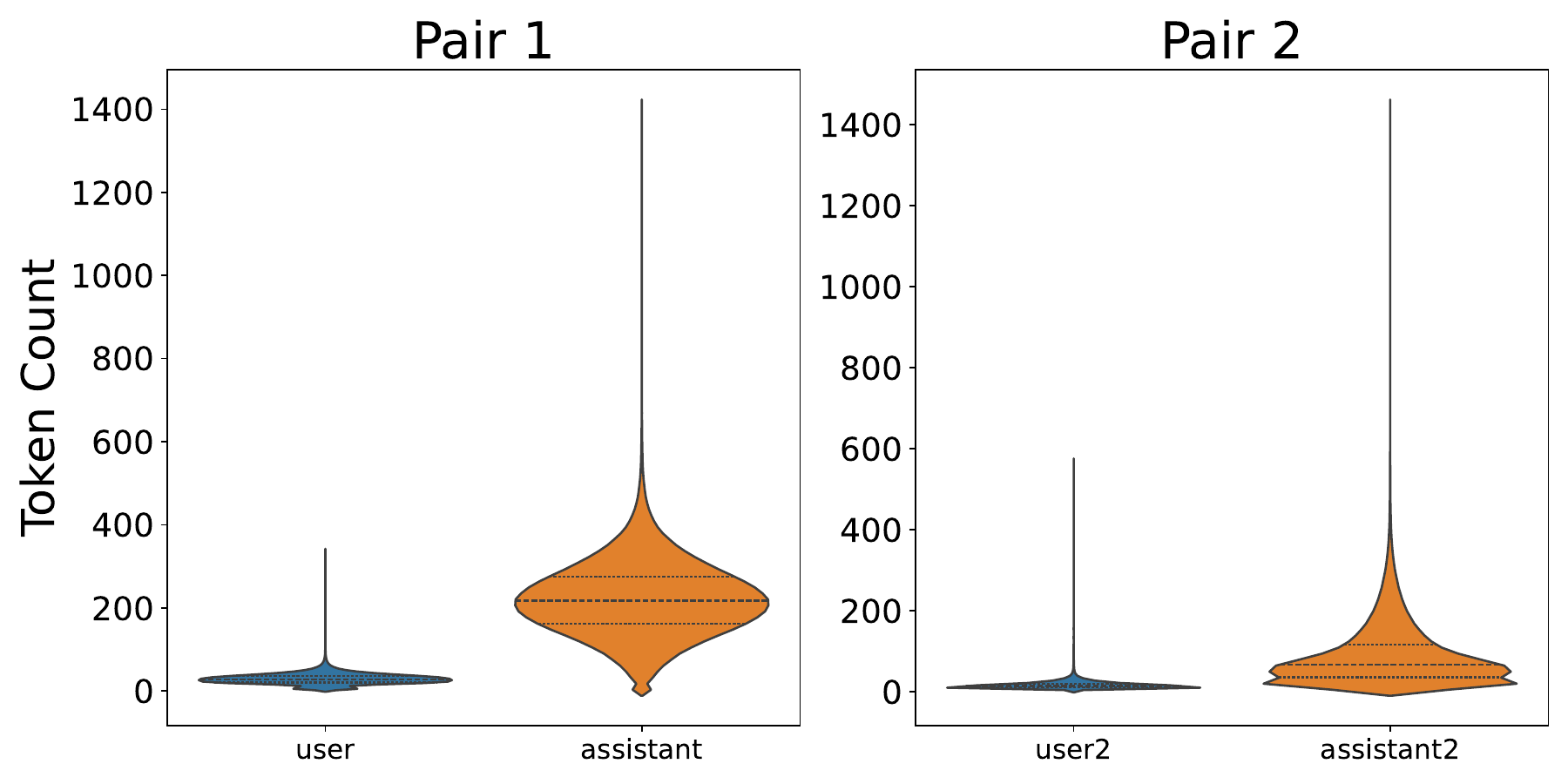}
    \caption{White-space token count comparison for Questions and Answers in Writing multi-turn data split.}
    \label{fig:writing-tokens}
\end{figure}

\begin{figure}
    \centering
    \includegraphics[width=0.85\linewidth]{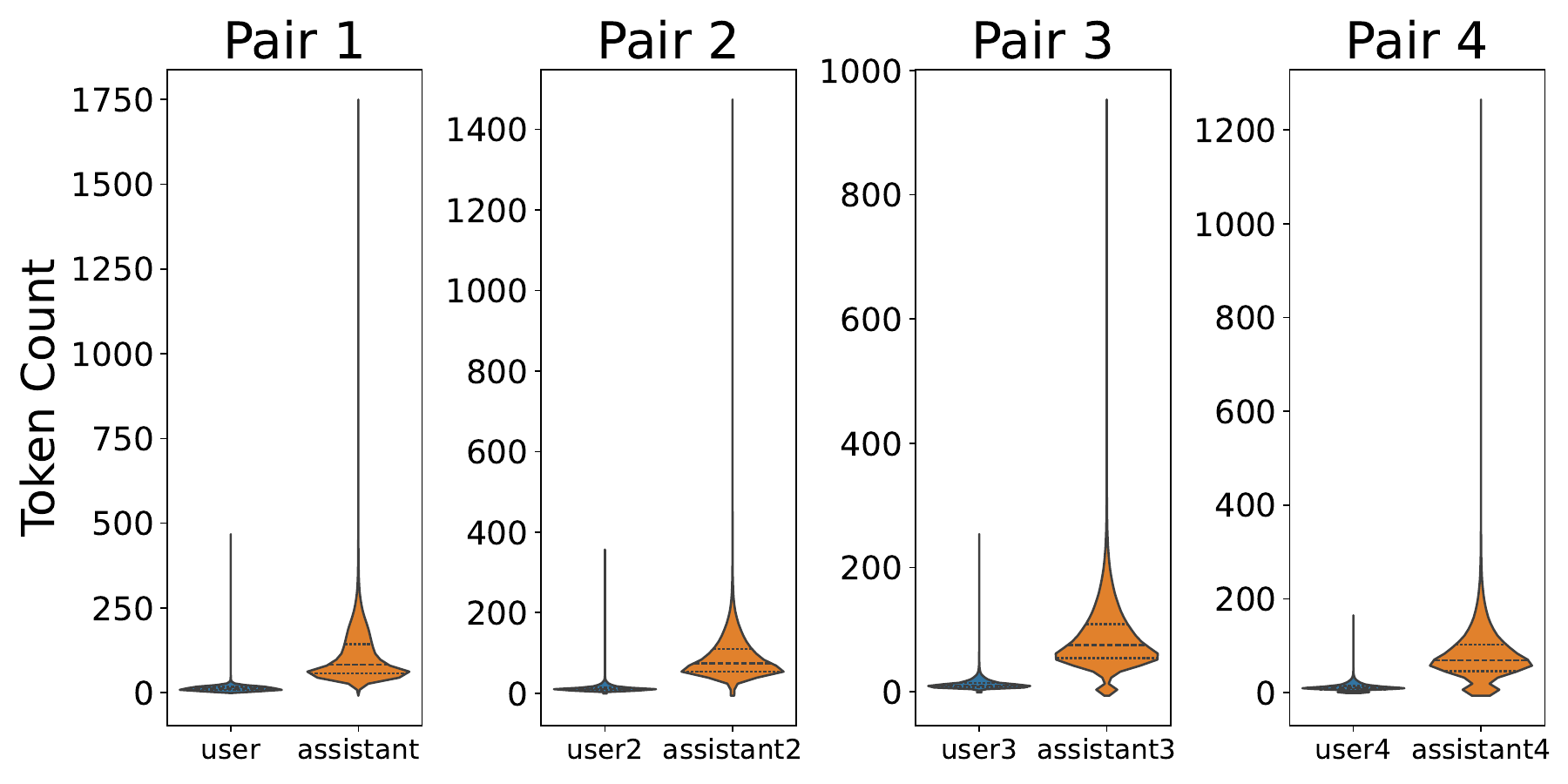}
    \caption{White-space token count comparison for Questions and Answers in Dialogue multi-turn data split.}
    \label{fig:dialog_tokens}
\end{figure}

In addition to the full token analysis in \Cref{fig:white-space_token_count_full_ylim}, we compute the token counts for each turn in multi-turn data. Please refer to \Cref{fig:writing-tokens} and \Cref{fig:dialog_tokens} for the Writing and Dialogue results.

\clearpage

\subsection{Topics Used in the Generation of the Instructions}\label{sec:topics}
The Academic, Multiple Choice, Dialogue, Translation and MMLU rely on the following topic lists when prompting the generation of pairs, with the later utilizing MMLU topics only.

Below is the list of topics that were used across Multiple Choice, Dialogue, and Translation splits.  This topic list was written by Gemini in response to the prompt ``Write a long list of topics you can answer questions about.''  The model was queried multiple times and the outputs were concatenated and deduplicated.

\textbf{Topics: }\texttt{Computer Science, Computer Programming, Python Programming, Java Programming, C++ Programming, Data structures and algorithms, Operating systems, Computer architecture, Networking, Artificial intelligence, Machine learning, Data Science, Data mining, Machine learning, Data visualization, Statistics, Data management, Data warehousing, Big data, Mathematics, Calculus, Linear algebra, Differential equations, Probability, Statistics, Real analysis, Complex analysis, Physics, Classical mechanics, Electromagnetism, Thermodynamics, Quantum mechanics, Special relativity, General relativity, Nuclear physics, Particle physics, Chemistry, General chemistry, Organic chemistry, Inorganic chemistry, Physical chemistry, Analytical chemistry, Biochemistry, Biology, Cell biology, Molecular biology, Genetics, Ecology, Evolution, Physiology, History, World history, American history, European history, Asian history, African history, Latin American history, Literature, English literature, American literature, European literature, Asian literature, African literature, Latin American literature, Art, Painting, Sculpture, Architecture, Music, Dance, Theater, Philosophy, Metaphysics, Epistemology, Ethics, Political philosophy, Aesthetics, Economics, Microeconomics, Macroeconomics, International trade, Public finance, Monetary policy, Economic development, Psychology, Cognitive psychology, Social psychology, Developmental psychology, Clinical psychology, Abnormal psychology, Sociology, Social stratification, Social inequality, Social mobility, Race and ethnicity, Gender and sexuality, Crime and deviance, Political Science, Comparative politics, International relations, American politics, Public policy, Political theory, Anthropology, Cultural anthropology, Social anthropology, Linguistic anthropology, Archaeological anthropology, Biological anthropology, Environmental Science, Ecology, Environmental chemistry, Environmental physics, Environmental biology, Environmental engineering, Environmental policy, Engineering, Mechanical engineering, Civil engineering, Electrical engineering, Chemical engineering, Computer engineering, Materials science engineering, Business, Accounting, Finance, Marketing, Management, Operations research, Information systems, Law, Constitutional law, Criminal law, Civil law, International law, Environmental law, Business law, Medicine, Anatomy, Physiology, Biochemistry, Microbiology, Pharmacology, Pathology, Surgery, Pediatrics, Psychiatry}

Below is a list of topics that were used to create the MMLU split:

\textbf{MMLU Topics: }\texttt{Abstract Algebra, Anatomy, Astronomy, Business Ethics, Clinical Knowledge, College Biology, College Chemistry, College Computer Science, College Mathematics, College Medicine, College Physics, Computer Security, Conceptual Physics, Econometrics, Electrical Engineering, Elementary Mathematics, Formal Logic, Global Facts, High School Biology, High School Chemistry, High School Computer Science, High School European History, High School Geography, High School Government And Politics, High School Macroeconomics, High School Mathematics, High School Microeconomics, High School Physics, High School Psychology, High School Statistics, High School US History, High School World History, Human Aging, Human Sexuality, International Law, Jurisprudence, Logical Fallacies, Machine Learning, Management, Marketing, Medical Genetics, Miscellaneous, Moral Disputes, Moral Scenarios, Nutrition, Philosophy, Prehistory, Professional Accounting, Professional Law, Professional Medicine, Professional Psychology, Public Relations, Security Studies, Sociology, US Foreign Policy, Virology, World Religions}

Below is the list of topics, programming language, libraries and markup languages used whilst creating the coding split.

\textbf{Coding Topics: }\texttt{Data Structures, Algorithms, Object-Oriented Programming, File Handling, Database Programming, Networking, Operating Systems, Web Development, Machine Learning, Data Analysis And Visualization, Software Testing, Software Development, Software Deployment, Cloud Computing, Blockchain, Machine Learning} 

\textbf{Coding Languages: }\texttt{Python, Java, C, C++, JavaScript, PHP, R, Swift, Go, Ruby, Kotlin, Scala, Rust, Haskell, Elixir, Julia, Lua, Groovy, Objective-C, Perl, Fortran, Visual Basic, MATLAB, SAS, COBOL}

\textbf{Libraries: }\texttt{Altair, Ansible, BeautifulSoup, Bokeh, Bottle, CatBoost, Chef, CherryPy, Click, Cocos2d, DearPyGUI, Django, Django ORM, Fabric, FastAPI, Fire, Flask, Flask-SQLAlchemy, Folium, Gensim, Godot Engine, Hugging Face, Hugging Face Transformers, Inquirer, Keras, Kivy, LightGBM, Matplotlib, MechanicalSoup, MongoEngine, Natural Language Toolkit (NLTK), NumPy, Panda3D, Pandas, Paramiko, Pattern, Plotly, Plotnine, Pony ORM, Puppet, PyGtk, PyMC3, PyOpenGL, PyParsing, PyQt, PySide, PySimpleGUI, PyTorch, PyTorch Lightning, Pygal, Pygame, Pyglet, Pyjion, Pyramid, Qtile, Quart, Ren'Py, Requests, RoboBrowser, SQLAlchemy, SQLModel, SaltStack, Scapy, SciPy, Scrapy, Seaborn, Selenium, Socket, Starlette, SymPy, TensorFlow, TextBlob, Tkinter, Tortoise ORM, TurboGears, Twisted, Urllib2, Web2py, XGBoost, aiobotocore, aiohttp, aiosignal, anyio, asn1crypto, async-timeout, asyncio, attrs, awscli, azure-core, beautifulsoup4, boto3, botocore, cachetools, certifi, cffi, charset-normalizer, click, colorama, coverage, cryptography, decorator, deprecated, distlib, docutils, et-xmlfile, exceptiongroup, filelock, flask, frozenlist, fsspec, gevent, ggplot, google-api-core, google-auth, google-cloud-core, google-cloud-storage, googleapis-common-protos, greenlet, grequests, grpcio, grpcio-status, idna, importlib-metadata, importlib-resources, iniconfig, isodate, jellyfish, jinja2, jmespath, jsonschema, langdetect, lxml, markupsafe, missingno, more-itertools, msgpack, multidict, nltk.translate, numpy, oauthlib, openpyxl, packaging, pandas, peewee, pillow, pip, platformdirs, plotly express, pluggy, protobuf, psutil, pyarrow, pyasn1, pyasn1-modules, pycparser, pydantic, pydantic-core, pygments, pyjwt, pymongo, pyopenssl, pyparsing, pytest, python-dateutil, pytz, pyyaml, requests, requests-oauthlib, rsa, s3fs, s3transfer, scikit-image, scikit-learn, scipy, setuptools, six, sniffio, soupsieve, spaCy, sqlalchemy, statsmodels, tomli, tomlkit, tqdm, typing-extensions, tzdata, urllib3, virtualenv, websocket-client, websockets, werkzeug, wheel, wrapt, wxPython, yarl, zipp}

\textbf{Markup Languages: }\texttt{PL/SQL, SQL, HTML, HTML, CSS, XML, JSON, YAML, Markdown, Latex}

Finally, we provide the document types list used in the generation of the Writing split. 

\textbf{Document Types: }\texttt{licensing report, offer letter, telemarketing script, training manual, privacy policy, job application, business proposal, notice, anniversary card, audiobook, human resources manual, friendship letter, statute of frauds, best evidence, project proposal, divorce papers, power of attorney, deposition notice, form, how-to guide, podcast, libretto, catalog, non-disclosure agreement (nda), loan application, return label, work of art, inquiry letter, zoning regulations, white paper, opinion editorial, vision statement, e-mail, worksheet, pen pal message, press release, building code, letter of intent, powerpoint presentation, conference proceedings, exhibit list, standard operating procedure (SOP), toast, disciplinary action, tax return, invitation, field report, chronology, ballad, job posting, guide, newsletter article, newspaper article, verdict form, zoning variance request, discovery plan, gratitude journal, guidebook, label, reformation, fundraising letter, standard operating procedure, list, complaint, marketing plan, federal register, membership card, purchase order, tax form, unpublished opinion, operating manual, quality assurance plan, literary analysis, music review, romance novel, craft project, feasibility analysis, declaration, degree, wish list, wills and testaments, shareholders' agreement, friendship card, marketing report, thank-you letter, legal encyclopedia, travel journal, apology letter, county ordinance, agreement, record, love card, problem statement, scientific paper, visa, descriptive essay, dream journal, conference paper, menu, law review article, business card, discovery request, adoption papers, greeting card, brief, master's thesis, writ of certiorari, timeline, financial plan, progress report, customer satisfaction survey, application letter, shipping order, award, mortgage document, hymn, fairy tale, licensing agreement, dissenting opinion, employee id card, health code, notice of termination, daily log, textbook, history, confidentiality agreement, comparison report, indemnity, journal entry, order form, warning, personal essay, discussion board post, per curiam opinion, inspection report, description, congressional record, technical report, policy brief, hypothesis, specific performance, opera, marriage license application, permit, poem, information sheet, interrogatories, book, zen koan, operations report, blog entry, thank you letter, medical record, meditation, merger clause, article, illustration, economic impact statement, audit report, internship application, acquisition agreement, balance sheet, social security card, problem-solution essay, town ordinance, independent contractor agreement, press kit, investigative report, law textbook, father's day card, brochure, travel itinerary, insurance application, management letter, poster, letter of reference, employee handbook, trademark report, fax, legal treatise, risk assessment, drama script, sales report, grant proposal, receipt, court rule, subrogation, patent notice, commercial, resume, database, performance review, event proposal, registration, advertising copy, social responsibility report, proclamation, music video script, cancellation, to-do list, operations plan, diploma, check, historical novel, business letter, market research report, comic book, contract, motion picture script, non-fiction book, registration report, software application, rescission, competency, public service announcement, statutory code, annotated bibliography, interview transcript, script, flyer, fiction novel, direct mail, event planning guide, statement of cash flows, academic journal, lawsuit, employment agreement, legal brief, lease agreement, comic strip, research paper, iou, curriculum vitae/resume, gift card, bucket list, award certificate, operating procedure, billboard, medical directive, amicus curiae brief, distribution agreement, diary entry, last will and testament, fanfiction, graduation card, social impact statement, award nomination, home warranty, arrest warrant, lease, credit application, manual of style, promissory note, packing list, credit agreement, prayer, will and testament, meeting agenda, song lyrics, journal article, customer service report, user manual, movie script, lyric, book proposal, job description, severance agreement, music, fan letter, cease and desist letter, screenplay, trade journal, debriefing report, warranty, procedure, weekly report, research and development plan, travel brochure, joint venture agreement, user agreement, statement, evaluation, anonymous letter, deposition, guideline, dissertation, mission statement, sales letter, retirement plan, prayer journal, flowchart, book chapter, legal dictionary, work of poetry, affirmation, real estate listing, appellate brief, immigration form, work breakdown structure, handwritten letter, instructional manual, wiki article, encyclopedia article, thesis or dissertation, restitution, autobiography, zoning permit, bylaws, answer, policy, order, review, thesis statement, thesis, popular magazine, public notice, fire code, application form, historical fiction, sermon, scientific report, open letter, trust, policy manual, creative nonfiction, feasibility report, problem-solving report, bid, business case, birthday card, birth certificate, human resources policy, copyright report, subpoena, policy and procedure manual, financial report, inventory, research proposal, speech, handbook, mortgage agreement, childcare agreement, performance evaluation, casebook, children's book, food journal, order to show cause, technical brief, explanation, christmas card, sympathy card, budget, risk management report, discussion, webinar, certificate, meeting minutes, opinion, public relations plan, comment, money order, yearbook, medical consent form, letter of resignation, disclaimer, paragraph, user's manual, blog, license, quarterly business review, auto/biography, scope of work, investment agreement, affirmation list, letter, divorce decree, summary, curriculum vitae (cv), zoning ordinance, affidavit, recipe, home budget, law, verdict, letter of apology, village ordinance, guarantee, design document, investment plan, interview, daily report, informal letter, movie review, opinion piece, pitch deck, marriage license, spreadsheet, injunction, search warrant, city ordinance, parol evidence rule, memoir, court order, maintenance manual, explanatory letter, appeal, memo, regulation, pest analysis, mystery novel, copyright application, governance report, bill, advice column, magazine article, independent study proposal, mortgage, legend, online advertisement, get well card, thank you card, corporate social responsibility report, youtube video script, literary magazine, yelp review, audio recording, critique, monograph, eulogy, request for proposal, cross-examination, user guide, waiver, history book, debit card, articles of incorporation, curriculum vitae, fiction, trial brief, email, essay, quality assurance report, sales order, story, impeachment, legal contract, feasibility study, newspaper, advertisement, consulting report, writ, recommendation report, travel guide, safety code, résumé, compliance report, monthly report, synopsis, marketing brochure, pen pal letter, anonymous card, loan agreement, homeowner's association rules, television advertisement, bill of sale, play, public service announcement (psa), memorandum opinion, confirmation letter, demand letter, reflective essay, marriage certificate, data report, mobile app, shipping label, conference proceeding, rating, goal list, novation, case management plan, fact sheet, demonstration, project report, SWOT analysis, character sketch, radio advertisement, license application, consent form, social media post, bullet journal, letter of recommendation, pleading, infomercial, information technology report, inventory list, medical report, thank-you note, visa application, survey, drama, quarterly report, comedy sketch, travelogue, petition, newsletter, insurance policy, literature review, ordinance, certification report, diary, manifesto, motion, post-trial brief, company policy manual, minutes, abstract, technical paper, announcement, memorandum of understanding, chain card, statement of work, exposé, product description, todo list, constitution, text message marketing campaign, zoning board of appeals decision, opening statement, course syllabus, complaint letter, trademark notice, non-compete agreement, probate document, portfolio, human resources plan, cookbook, financial aid application, invitation letter, voter registration card, game, dental record, direct examination, letter of credit, love letter, dictionary, petition for review, jury instructions, legal document, analytical report, accord and satisfaction, leaflet, administrative code, product manual, gift certificate, threat card, newspaper article, default judgment, satire, commentary, mother's day card, zoning code, humorous story, employment application, closing argument, environmental code, invoice, termination letter, definition, student id card, quality assurance manual, editorial, evaluation form, workshop proposal, recommendation letter, contribution, meme, letter of agreement, chronicle, development plan, fable, analysis, direct mail marketing campaign, temporary restraining order, character reference, short story, concurring opinion, self-help book, software documentation, play script, profile, statute of limitations, storyboard, transcript, status report, credit card application, personal statement, quiz, obituary, project plan, living will, engineering report, diy project, informational article, blueprint, historical document, merger agreement, email marketing campaign, index, damages, death certificate, blog post, paranormal story, environmental report, work order, manual, academic paper, summary judgment, medicare card, patent, income statement, fan card, discussion paper, claim, investment prospectus, sustainability report, instruction manual, warning letter, grocery list, accounting report, citation, covenant, safety manual, annual report, cv/resume, speech transcript, song, novel, gantt chart, congratulations card, bench warrant, accreditation report, privilege, mystery, human resources report, scholarship application, request for production of documents, evaluation report, deed, thesis, school transcript, report, will, legal report, valentine's day card, statistical report, photo essay, interrogatory, lyrics, film script, apology card, regulations, criticism, comparative analysis, presentation, judgment, technical manual, online article, video podcast, book review, rehabilitation, biography, zine, lien, release form, terms of service, employee satisfaction survey, pert chart, questionnaire, magazine, outline, hearsay, financial analysis, directive, reference guide, passport, manuscript, video, humor, cover letter, vision board, packing slip, financial statement, statement of retained earnings, illustrated story, plan, tabloid, patent application, relevance, informational brochure, handout, itinerary, thank you note, coupon, environmental impact statement, estoppel, query letter, statement of purpose, encyclopedia, incident report, adventure story, business plan, online course, non-disclosure agreement, guestbook entry, case study, cashier's check, news article, grant application, creative writing, expense report, formal letter, state statute, thank-you card, company profile, summons, gratitude list, policy statement, reflection, infographic, autobiographical essay, copyright notice, pamphlet, strategic plan, application, partnership agreement, decision report, specification, condolence letter, bibliography, journal, graphic novel, agenda, integration clause, reference letter, redirect examination, conclusion, product review, wedding invitation, organizational chart, proposal, study guide, campaign speech, homily, art project, information technology plan.}

\end{document}



\centering

{\large \textbf{GenQA: An Instruction Dataset of LLM Generated Questions and Answers}} \\
\vspace{0.5em}{\large Supplementary Material} \\
\vspace{1.5em}

\raggedright

\textbf{Author Statement.} We bear the responsibility for maintenance, licensing, and distribution of our dataset.

\textbf{License.} GenQA will be released under the \textit{Creative Commons Attribution Non-Commercial Share Alike 4.0} license.

\textbf{Hosting URLs.} 

(For peer review purposes, the dataset is hosted under anonymous links. Public versions of the same artifacts will be released with the announcement and public release of the associated manuscript.)

The GenQA data is provided in three formats: as $9$ individual splits, as a single dataset containing all splits concatenated together, and in a re-balanced form where some large splits are down sampled before concatenation for training purposes. For ease of use, since many of the splits in GenQA are multi-turn, they are serialized in a unified format where the set of user and assistant turns comprising single sample are represented as a list of json objects.

\paragraph{GenQA Splits} The GenQA splits are \texttt{academic}, \texttt{mmlu}, \texttt{multiple\_choice}, \texttt{writing}, \texttt{task}, \texttt{code}, \texttt{math}, \texttt{dialog}, and \texttt{general}. They can be individually accessed using the following link template.

Dataset \url{https://huggingface.co/datasets/anaonymous-aad/GenQA_<split>}

Croissant metadata \url{https://huggingface.co/api/datasets/anaonymous-aad/GenQA_<split>/croissant}

\paragraph{Raw GenQA} This is the complete version of the GenQA dataset containing the result of concatenating all $9$ splits without any re-sampling.

Dataset \url{https://huggingface.co/datasets/anaonymous-aad/GenQA_Full}

Croissant metadata \url{https://huggingface.co/api/datasets/anaonymous-aad/GenQA_Full/croissant}

\paragraph{GenQA} The re-balanced version from the empirical evaluations in the manuscript. 

Dataset \url{https://huggingface.co/datasets/anaonymous-aad/GenQA}

Croissant metadata \url{https://huggingface.co/api/datasets/anaonymous-aad/GenQA/croissant}

\paragraph{Model checkpoints} The model checkpoints produced during the empirical evaluation described in the associated manuscript will be made publicly available when it is publicly released. Upon request, access to the checkpoints can be granted as part of the review process.

\section*{GenQA Datasheet}
\label{sec:datasheet}
\dssectionheader{Motivation}

\dsquestionex{For what purpose was the dataset created?}{Was there a specific task in mind? Was there a specific gap that needed to be filled? Please provide a description.}

\dsanswer{GenQA was created to demonstrate the effectiveness of autonomously written instruction datasets, and to produce a research instruction dataset of size comparable to commercial instruction sets.}

\dsquestion{Who created this dataset (e.g., which team, research group) and on behalf of which entity (e.g., company, institution, organization)?}

\dsanswer{The dataset was created by researchers from the University of Maryland, College Park.}

\bigskip
\dssectionheader{Composition}

\dsquestionex{What do the instances that comprise the dataset represent (e.g., documents, photos, people, countries)?}{ Are there multiple types of instances (e.g., movies, users, and ratings; people and interactions between them; nodes and edges)? Please provide a description.}

\dsanswer{Each instance contains simulated conversations between a human user and an LLM virtual assistant.  Many splits contain instances with multiple ``turns'' in which the human and machine alternate questions and responses.}

\dsquestion{How many instances are there in total (of each type, if appropriate)?}

\dsanswer{There are approximately $11$M instances in the raw concatenated version containing all $9$ splits. The re-sampled version contains approximately $6.5$M instances.}

\dsquestionex{Does the dataset contain all possible instances or is it a sample (not necessarily random) of instances from a larger set?}{ If the dataset is a sample, then what is the larger set? Is the sample representative of the larger set (e.g., geographic coverage)? If so, please describe how this representativeness was validated/verified. If it is not representative of the larger set, please describe why not (e.g., to cover a more diverse range of instances, because instances were withheld or unavailable).}

\dsanswer{The dataset is a representative sample of possible questions that could be generated using the selected prompts.}

\dsquestionex{What data does each instance consist of? “Raw” data (e.g., unprocessed text or images) or features?}{In either case, please provide a description.}

\dsanswer{Each sample contains Unicode formatted text representing a conversation between a user and an assistant. A unified format is used for both single and multi-turn conversations.}

\dsquestionex{Is there a label or target associated with each instance?}{If so, please provide a description.}

\dsanswer{There is no explicit target, but most often the ``assistant'' outputs in a given conversation will be used as targets for supervised instruction tuning.}

\dsquestionex{Is any information missing from individual instances?}{If so, please provide a description, explaining why this information is missing (e.g. because it was unavailable). This does not include intentionally removed information but might include, e.g., redacted text.}

\dsanswer{N/A}

\dsquestionex{Are relationships between individual instances made explicit (e.g., users’ movie ratings, social network links)?}{If so, please describe how these relationships are made explicit.}

\dsanswer{Yes, instances produced by similar methods appear in the same ``split.''}

\dsquestionex{Are there recommended data splits (e.g., training, development/validation, testing)?}{If so, please provide a description of these splits, explaining the rationale behind them.}

\dsanswer{Yes. The associated manuscript contains a detailed explanation of each split.  Each split pertains to a different range of topics.}

\dsquestionex{Are there any errors, sources of noise, or redundancies in the dataset?}{If so, please provide a description.}

\dsanswer{The dataset may contain factual inaccuracies and was not manually checked for factual correctness.}

\dsquestionex{Is the dataset self-contained, or does it link to or otherwise rely on external resources (e.g., websites, tweets, other datasets)?}{If it links to or relies on external resources, a) are there guarantees that they will exist, and remain constant, over time; b) are there official archival versions of the complete dataset (i.e., including the external resources as they existed at the time the dataset was created); c) are there any restrictions (e.g., licenses, fees) associated with any of the external resources that might apply to a future user? Please provide descriptions of all external resources and any restrictions associated with them, as well as links or other access points, as appropriate.}

\dsanswer{The dataset is a self contained artifact that will be publicly released and made freely available online.}

\dsquestionex{Does the dataset contain data that might be considered confidential (e.g., data that is protected by legal privilege or by doctor-patient confidentiality, data that includes the content of individuals non-public communications)?}{If so, please provide a description.}

\dsanswer{No.}

\dsquestionex{Does the dataset contain data that, if viewed directly, might be offensive, insulting, threatening, or might otherwise cause anxiety?}{If so, please describe why.}

\dsanswer{Since the dataset was generated using the Gemini language model family, all data passed through the Gemini toxicity filters in the process, and so individual samples are unlikely to contain offensive material.}

\dsquestionex{Does the dataset relate to people?}{If not, you may skip the remaining questions in this section.}

\dsanswer{N/A}

\dsquestionex{Does the dataset identify any subpopulations (e.g., by age, gender)?}{If so, please describe how these subpopulations are identified and provide a description of their respective distributions within the dataset.}

\dsanswer{N/A}

\dsquestionex{Is it possible to identify individuals (i.e., one or more natural persons), either directly or indirectly (i.e., in combination with other data) from the dataset?}{If so, please describe how.}

\dsanswer{N/A}

\dsquestionex{Does the dataset contain data that might be considered sensitive in any way (e.g., data that reveals racial or ethnic origins, sexual orientations, religious beliefs, political opinions or union memberships, or locations; financial or health data; biometric or genetic data; forms of government identification, such as social security numbers; criminal history)?}{If so, please provide a description.}

\dsanswer{N/A}

\bigskip
\dssectionheader{Collection Process}

\dsquestionex{How was the data associated with each instance acquired?}{Was the data directly observable (e.g., raw text, movie ratings), reported by subjects (e.g., survey responses), or indirectly inferred/derived from other data (e.g., part-of-speech tags, model-based guesses for age or language)? If data was reported by subjects or indirectly inferred/derived from other data, was the data validated/verified? If so, please describe how.}

\dsanswer{Questions were written by the Gemini Language model in response to a human-written prompt.}

\dsquestionex{What mechanisms or procedures were used to collect the data (e.g., hardware apparatus or sensor, manual human curation, software program, software API)?}{How were these mechanisms or procedures validated?}

\dsanswer{The dataset was synthetically generated using advanced commercial LLMs that were accessed via API calls.}

\dsquestion{Who was involved in the data collection process (e.g., students, crowdworkers, contractors) and how were they compensated (e.g., how much were crowdworkers paid)?}

\dsanswer{Datasets were collected by university faculty and graduate students.}

\dsquestionex{Over what timeframe was the data collected? Does this timeframe match the creation timeframe of the data associated with the instances (e.g., recent crawl of old news articles)?}{If not, please describe the timeframe in which the data associated with the instances was created.}

\dsanswer{The dataset was collected from Feburary 2024 to June 2024.}

\dsquestionex{Were any ethical review processes conducted (e.g., by an institutional review board)?}{If so, please provide a description of these review processes, including the outcomes, as well as a link or other access point to any supporting documentation.}

\dsanswer{No.}

\dsquestionex{Does the dataset relate to people?}{If not, you may skip the remaining questions in this section.}

\dsanswer{N/A}

\dsquestion{Did you collect the data from the individuals in question directly, or obtain it via third parties or other sources (e.g., websites)?}

\dsanswer{N/A}

\dsquestionex{Were the individuals in question notified about the data collection?}{If so, please describe (or show with screenshots or other information) how notice was provided, and provide a link or other access point to, or otherwise reproduce, the exact language of the notification itself.}

\dsanswer{N/A}

\dsquestionex{Did the individuals in question consent to the collection and use of their data?}{If so, please describe (or show with screenshots or other information) how consent was requested and provided, and provide a link or other access point to, or otherwise reproduce, the exact language to which the individuals consented.}

\dsanswer{N/A}

\dsquestionex{If consent was obtained, were the consenting individuals provided with a mechanism to revoke their consent in the future or for certain uses?}{If so, please provide a description, as well as a link or other access point to the mechanism (if appropriate).}

\dsanswer{N/A}

\dsquestionex{Has an analysis of the potential impact of the dataset and its use on data subjects (e.g., a data protection impact analysis) been conducted?}{If so, please provide a description of this analysis, including the outcomes, as well as a link or other access point to any supporting documentation.}

\dsanswer{N/A}

\bigskip
\dssectionheader{Preprocessing/cleaning/labeling}

\dsquestionex{Was any preprocessing/cleaning/labeling of the data done (e.g., discretization or bucketing, tokenization, part-of-speech tagging, SIFT feature extraction, removal of instances, processing of missing values)?}{If so, please provide a description. If not, you may skip the remainder of the questions in this section.}

\dsanswer{Exact deduplication was done on the first two sentences of each question.}

\dsquestionex{Was the “raw” data saved in addition to the preprocessed/cleaned/labeled data (e.g., to support unanticipated future uses)?}{If so, please provide a link or other access point to the “raw” data.}

\dsanswer{No.}

\dsquestionex{Is the software used to preprocess/clean/label the instances available?}{If so, please provide a link or other access point.}

\dsanswer{N/A.}

\bigskip
\dssectionheader{Distribution}

\dsquestionex{Will the dataset be distributed to third parties outside of the entity (e.g., company, institution, organization) on behalf of which the dataset was created?}{If so, please provide a description.}

\dsanswer{The dataset will be made publicly available upon announcement and publication of the associated manuscript.}

\dsquestionex{How will the dataset be distributed (e.g., tarball on website, API, GitHub)}{Does the dataset have a digital object identifier (DOI)?}

\dsanswer{The dataset will be distributed via the widely popular machine learning platform Hugging Face and will be downloadable using their \texttt{huggingface\_hub} Python client library.}

\dsquestion{When will the dataset be distributed?}

\dsanswer{The dataset will be made publicly available upon announcement and publication of the associated manuscript.}

\dsquestionex{Will the dataset be distributed under a copyright or other intellectual property (IP) license, and/or under applicable terms of use (ToU)?}{If so, please describe this license and/or ToU, and provide a link or other access point to, or otherwise reproduce, any relevant licensing terms or ToU, as well as any fees associated with these restrictions.}

\dsanswer{The dataset will be released under the CC BY-NC-SA license.}

\dsquestionex{Have any third parties imposed IP-based or other restrictions on the data associated with the instances?}{If so, please describe these restrictions, and provide a link or other access point to, or otherwise reproduce, any relevant licensing terms, as well as any fees associated with these restrictions.}

\dsanswer{No.}

\dsquestionex{Do any export controls or other regulatory restrictions apply to the dataset or to individual instances?}{If so, please describe these restrictions, and provide a link or other access point to, or otherwise reproduce, any supporting documentation.}

\dsanswer{No.}

\bigskip
\dssectionheader{Maintenance}

\dsquestion{Who will be supporting/hosting/maintaining the dataset?}

\dsanswer{The dataset will be hosted on Hugging Face, a widely popular platform for sharing machine learning datasets. The authors will maintain the dataset to address any errata, future iterations, and other updates.}

\dsquestion{How can the owner/curator/manager of the dataset be contacted (e.g., email address)?}

\dsanswer{Author contact information is omitted for review purposes but an email will be provided in the associated manuscript upon publication.}

\dsquestionex{Is there an erratum?}{If so, please provide a link or other access point.}

\dsanswer{Currently, there is no erratum.}

\dsquestionex{Will the dataset be updated (e.g., to correct labeling errors, add new instances, delete instances)?}{If so, please describe how often, by whom, and how updates will be communicated to users (e.g., mailing list, GitHub)?}

\dsanswer{We will update the dataset if any errors are identified by the authors. The dataset will be updated via Hugging Face, and users will be able to access the version history.}

\dsquestionex{If the dataset relates to people, are there applicable limits on the retention of the data associated with the instances (e.g., were individuals in question told that their data would be retained for a fixed period of time and then deleted)?}{If so, please describe these limits and explain how they will be enforced.}

\dsanswer{The dataset does not contain any information or responses collected from the general public, crowdworkers, etc.}

\dsquestionex{Will older versions of the dataset continue to be supported/hosted/maintained?}{If so, please describe how. If not, please describe how its obsolescence will be communicated to users.}

\dsanswer{Yes, since the dataset will be hosted on Hugging Face, users will be able to access any older version of the datasets.}

\dsquestionex{If others want to extend/augment/build on/contribute to the dataset, is there a mechanism for them to do so?}{If so, please provide a description. Will these contributions be validated/verified? If so, please describe how. If not, why not? Is there a process for communicating/distributing these contributions to other users? If so, please provide a description.}

\dsanswer{To keep results reproducible, we do not intend to significantly alter the dataset unless major errors are identified.  Future extensions of the dataset will be re-released as a new ``GenQA version 2'' with an updated description and model card.}